\documentclass{article}

\usepackage{arxiv}
\usepackage[utf8]{inputenc}
\usepackage[T1]{fontenc}
\usepackage{hyperref}
\usepackage{url}
\usepackage{booktabs}
\usepackage{amsfonts}
\usepackage{amsmath}
\usepackage{microtype}
\usepackage{graphicx}
\usepackage{natbib}
\usepackage{doi}
\usepackage{xcolor}
\usepackage{xspace}
\usepackage{tikz}
\usetikzlibrary{arrows.meta,positioning}
\xspaceaddexceptions{\%}

\newcommand{\NTestClips}{800\xspace}
\newcommand{\NPoolClips}{799\xspace}
\newcommand{\NAItems}{120\xspace}
\newcommand{\NARepeats}{20\xspace}
\newcommand{\NAMinGap}{23\xspace}
\newcommand{\NBItems}{50\xspace}
\newcommand{\NBShared}{30\xspace}
\newcommand{\NBNew}{20\xspace}
\newcommand{\NDistinctClips}{120\xspace}
\newcommand{\NAClassMin}{3\xspace}
\newcommand{\NAGapDesign}{20\xspace}
\newcommand{\NBNewClassMin}{2\xspace}

\newcommand{\NDistinctAbuse}{5\xspace}

\newcommand{\NAHours}{6.0\xspace}
\newcommand{\NAMedianMinutes}{2.6\xspace}

\newcommand{\NAMinWatch}{84\xspace}
\newcommand{\NBHours}{2.3\xspace}
\newcommand{\NBMedianMinutes}{1.8\xspace}

\newcommand{\NBMinWatch}{70\xspace}
\newcommand{\NWatchThreshold}{80\xspace}

\newcommand{\NAClips}{100\xspace}
\newcommand{\NADecided}{97\xspace}
\newcommand{\NAUnsure}{3\xspace}
\newcommand{\NAKappa}{0.95\xspace}
\newcommand{\NAKappaLo}{0.88\xspace}
\newcommand{\NAKappaHi}{1.00\xspace}
\newcommand{\NAAcc}{97.9\xspace}

\newcommand{\NAKappaUnsureYes}{0.91\xspace}
\newcommand{\NAKappaUnsureNo}{0.93\xspace}

\newcommand{\NAYesNormal}{2\xspace}

\newcommand{\NBDecided}{48\xspace}
\newcommand{\NBUnsure}{2\xspace}

\newcommand{\NSelfKappa}{1.00\xspace}
\newcommand{\NSelfRecognised}{3\xspace}
\newcommand{\NABShared}{30\xspace}
\newcommand{\NABSame}{29\xspace}
\newcommand{\NABKappa}{0.93\xspace}
\newcommand{\NABKappaLo}{0.75\xspace}
\newcommand{\NABKappaHi}{1.00\xspace}
\newcommand{\NABKappaBinary}{0.91\xspace}
\newcommand{\NABBinaryN}{28\xspace}
\newcommand{\NABBothUnsure}{2\xspace}

\newcommand{\NABBothYes}{20\xspace}
\newcommand{\NABIoUMedian}{0.78\xspace}
\newcommand{\NABIoUMean}{0.73\xspace}
\newcommand{\NABIoUMin}{0.18\xspace}
\newcommand{\NABIoUHalf}{16\xspace}
\newcommand{\NABVisualSame}{20\xspace}
\newcommand{\NABSoundSame}{19\xspace}
\newcommand{\NABSpeechSame}{17\xspace}
\newcommand{\NABCategoryShared}{19\xspace}
\newcommand{\NASegments}{171\xspace}
\newcommand{\NASegmentsPerYes}{2.6\xspace}
\newcommand{\NBSegments}{113\xspace}
\newcommand{\NBSegmentsPerYes}{3.3\xspace}

\newcommand{\NAudioMusicOnly}{33\xspace}
\newcommand{\NAudioNearSilent}{1\xspace}
\newcommand{\NAudioWithSound}{765\xspace}
\newcommand{\NAudioWithSoundShare}{95.7\xspace}
\newcommand{\NAudioViolentSound}{96.0\xspace}
\newcommand{\NAudioNormalSound}{95.3\xspace}
\newcommand{\NAudioFisherP}{0.72\xspace}
\newcommand{\NAAllYesClips}{65\xspace}
\newcommand{\NAAllSoundOrSpeech}{86.2\xspace}
\newcommand{\NAAllSound}{81.5\xspace}
\newcommand{\NAAllSpeech}{21.5\xspace}

\newcommand{\NAWithSoundYesClips}{62\xspace}
\newcommand{\NAWithSoundSoundOrSpeech}{85.5\xspace}

\newcommand{\NBAllYesClips}{34\xspace}
\newcommand{\NBAllSoundOrSpeech}{94.1\xspace}
\newcommand{\NBAllSound}{94.1\xspace}
\newcommand{\NBAllSpeech}{23.5\xspace}

\newcommand{\NBWithSoundYesClips}{33\xspace}
\newcommand{\NBWithSoundSoundOrSpeech}{93.9\xspace}

\newcommand{\NTagTicked}{164\xspace}
\newcommand{\NTagUnticked}{120\xspace}
\newcommand{\NTagHeardTicked}{44\xspace}
\newcommand{\NTagHeardUnticked}{12\xspace}
\newcommand{\NTagHeardLowTicked}{71\xspace}
\newcommand{\NTagHeardLowUnticked}{41\xspace}
\newcommand{\NTagSpeechTicked}{27\xspace}
\newcommand{\NTagSpeechHeardTicked}{85\xspace}
\newcommand{\NTagSpeechHeardUnticked}{74\xspace}

\newcommand{\NLitLavadAP}{62.0\xspace}
\newcommand{\NLitEventVadAP}{64.0\xspace}
\newcommand{\NLitQvadAP}{68.5\xspace}
\newcommand{\NLitPandaAP}{70.2\xspace}
\newcommand{\NLitVadClipAP}{84.5\xspace}
\newcommand{\NLitHolmesVauAP}{87.7\xspace}
\newcommand{\NSysClips}{799\xspace}
\newcommand{\NSysViolent}{499\xspace}
\newcommand{\NSysNormal}{300\xspace}
\newcommand{\NSysOneAUC}{0.997\xspace}
\newcommand{\NSysOneF}{0.968\xspace}

\newcommand{\NSysOneTokens}{9{,}395\xspace}

\newcommand{\NSysOneTsFrameAP}{0.573\xspace}

\newcommand{\NSysOneTsGroundP}{0.551\xspace}
\newcommand{\NSysOneTsGroundR}{0.803\xspace}
\newcommand{\NSysOneTsFvsA}{0.985\xspace}

\newcommand{\NSysTwoFvsA}{0.913\xspace}

\newcommand{\NSysThreeFvsA}{0.952\xspace}

\newcommand{\NSysThreeBGroundP}{0.658\xspace}

\newcommand{\NSysFourF}{0.967\xspace}
\newcommand{\NSysFourFrameAP}{0.538\xspace}
\newcommand{\NSysFourFrameAPNoSeg}{0.535\xspace}
\newcommand{\NSysFourGroundP}{0.627\xspace}
\newcommand{\NSysFourGroundR}{0.774\xspace}
\newcommand{\NSysFourFvsA}{0.961\xspace}

\newcommand{\NSysFourTokens}{3{,}255\xspace}

\newcommand{\NSysQwenTokens}{17{,}874\xspace}
\newcommand{\NSysThreeCalls}{2.4\xspace}
\newcommand{\NSysQwenCalls}{6.0\xspace}
\newcommand{\NToolCallsThree}{1{,}904\xspace}
\newcommand{\NToolLinksThree}{3\xspace}
\newcommand{\NDiffThreeMinusTwo}{+0.005\xspace}
\newcommand{\NDiffThreeMinusTwoCI}{[$-$0.011, +0.020]\xspace}

\newcommand{\NDiffThreeMinusFour}{$-$0.032\xspace}
\newcommand{\NDiffThreeMinusFourCI}{[$-$0.046, $-$0.019]\xspace}

\newcommand{\NDiffFourMinusOne}{$-$0.001\xspace}
\newcommand{\NDiffFourMinusOneCI}{[$-$0.014, +0.012]\xspace}

\newcommand{\NDiffFourMinusOneTs}{+0.001\xspace}
\newcommand{\NDiffFourMinusOneTsCI}{[$-$0.012, +0.014]\xspace}

\newcommand{\NDiffThreeBMinusThree}{$-$0.011\xspace}
\newcommand{\NDiffThreeBMinusThreeCI}{[$-$0.027, +0.005]\xspace}

\newcommand{\NDiffThreeEdges}{+0.005\xspace}
\newcommand{\NDiffThreeEdgesCI}{[0.000, +0.011]\xspace}

\newcommand{\NDiffThreeBEdges}{$-$0.013\xspace}
\newcommand{\NDiffThreeBEdgesCI}{[$-$0.024, $-$0.003]\xspace}

\newcommand{\NDiffThreeSound}{+0.013\xspace}
\newcommand{\NDiffThreeSoundCI}{[+0.003, +0.024]\xspace}

\newcommand{\NDiffThreeBSound}{$-$0.002\xspace}
\newcommand{\NDiffThreeBSoundCI}{[$-$0.012, +0.008]\xspace}

\newcommand{\NDiffFourSound}{+0.012\xspace}
\newcommand{\NDiffFourSoundCI}{[+0.004, +0.021]\xspace}

\newcommand{\NDiffFourSpeech}{+0.005\xspace}
\newcommand{\NDiffFourSpeechCI}{[$-$0.001, +0.011]\xspace}

\newcommand{\NDiffFourSoundSpeech}{+0.017\xspace}
\newcommand{\NDiffFourSoundSpeechCI}{[+0.007, +0.027]\xspace}

\newcommand{\NDiffQwenMinusThree}{+0.004\xspace}
\newcommand{\NDiffQwenMinusThreeCI}{[$-$0.011, +0.020]\xspace}

\newcommand{\NFaithThreeCited}{46\xspace}
\newcommand{\NFaithThreeRandom}{2.9\xspace}

\newcommand{\NFaithFourCited}{49\xspace}
\newcommand{\NFaithFourRandom}{0.6\xspace}

\newcommand{\NCostFour}{20\xspace}
\newcommand{\NCostOne}{6.2\xspace}
\newcommand{\NCostRatio}{3\xspace}
\newcommand{\NDevClips}{152\xspace}
\newcommand{\NDevUsed}{151\xspace}
\newcommand{\NDevNormal}{60\xspace}
\newcommand{\NDevViolent}{92\xspace}
\newcommand{\NDevThreeOneF}{0.893\xspace}

\newcommand{\NDevAttemptCF}{0.907\xspace}
\newcommand{\NDevAttemptDF}{0.896\xspace}
\newcommand{\NDevFourSummariesF}{0.926\xspace}
\newcommand{\NDevFourF}{0.946\xspace}
\newcommand{\NDevReproF}{0.907\xspace}
\newcommand{\NGraphNodes}{51{,}542\xspace}
\newcommand{\NGraphNodesMedian}{46\xspace}
\newcommand{\NGraphNSoundEvent}{1{,}476\xspace}
\newcommand{\NGraphNTranscriptSegment}{10{,}634\xspace}
\newcommand{\NGraphNSpeechTag}{1{,}067\xspace}
\newcommand{\NGraphNCaption}{24{,}808\xspace}
\newcommand{\NGraphNDetection}{13{,}557\xspace}
\newcommand{\NGraphESameEvent}{30{,}530\xspace}
\newcommand{\NGraphESupports}{14{,}009\xspace}
\newcommand{\NGraphEDerivedFrom}{1{,}067\xspace}
\newcommand{\NGraphEContradicts}{16\xspace}
\newcommand{\NGraphContraOld}{11{,}188\xspace}
\newcommand{\NCapNegOld}{91\xspace}
\newcommand{\NCapNegNew}{6\xspace}
\newcommand{\NThreatOld}{715\xspace}
\newcommand{\NThreatNew}{129\xspace}

\newcommand{\NRobThreeQwenAUC}{0.979\xspace}

\newcommand{\NRobTwoQwenUnparsed}{2\xspace}
\newcommand{\NRobThreeQwenUnparsed}{6\xspace}
\newcommand{\NRobParaThree}{+0.001\xspace}
\newcommand{\NRobParaThreeCI}{[$-$0.011, +0.014]\xspace}

\newcommand{\NRobParaFour}{+0.003\xspace}
\newcommand{\NRobParaFourCI}{[$-$0.004, +0.009]\xspace}

\newcommand{\NRobParaThreeMinusFour}{$-$0.034\xspace}
\newcommand{\NRobParaThreeMinusFourCI}{[$-$0.048, $-$0.021]\xspace}

\newcommand{\NRobQwenThreeMinusFour}{$-$0.023\xspace}
\newcommand{\NRobQwenThreeMinusFourCI}{[$-$0.037, $-$0.010]\xspace}

\newcommand{\NRobQwenThreeMinusTwo}{+0.048\xspace}
\newcommand{\NRobQwenThreeMinusTwoCI}{[+0.029, +0.068]\xspace}

\newcommand{\NRobQwenFourMinusGemma}{$-$0.005\xspace}
\newcommand{\NRobQwenFourMinusGemmaCI}{[$-$0.015, +0.005]\xspace}

\newcommand{\NRobQwenTwoMinusGemma}{$-$0.039\xspace}
\newcommand{\NRobQwenTwoMinusGemmaCI}{[$-$0.059, $-$0.020]\xspace}

\newcommand{\NFailClips}{40\xspace}
\newcommand{\NFailViolent}{27\xspace}
\newcommand{\NFailNormal}{13\xspace}
\newcommand{\NFailAllShown}{12\xspace}
\newcommand{\NFailPartShown}{11\xspace}
\newcommand{\NFailNeverShown}{4\xspace}
\newcommand{\NFailShownShareAll}{50\xspace}
\newcommand{\NFailToolCallsAll}{2.4\xspace}
\newcommand{\NFailEarlyStop}{10\xspace}
\newcommand{\NFailMisjudged}{9\xspace}
\newcommand{\NFailDetectorLure}{8\xspace}
\newcommand{\NFailLabelDebatable}{9\xspace}

\newcommand{\NExDuration}{17\xspace}
\newcommand{\NExItems}{14\xspace}

\renewcommand{\shorttitle}{Language-model agents for violent video review}
\title{\textbf{What Helps Language-Model Agents Review Violent Video?\\ A Controlled Comparison with a Human Evidence Reference}}

\author{
  Mohammed El Amine Mokhtari \quad
  Matei Mancas \quad
  Virginie Vandenbulcke \\[0.8em]
  University of Mons
}

\date{}

\begin{document}
\maketitle

\begin{abstract}
Systems that flag violent video for a human operator can now be built from language models without training, by combining perception models and letting a language model decide. We ask how the observations of the perception models should reach the decision, and which evidence matters. On the XD-Violence test set (\NSysClips clips) we fix one set of perception agents (video description, object detection, sound tagging, speech recognition and speech tagging) and one reasoning model, and compare three interfaces: messages between agents, a typed evidence graph explored with tools, and flat text read in one pass; and a single audio-visual model given the raw clip. Two annotators provide a human reference that records the decision, the incident segments and whether they relied on sight, sound or speech. Flat text matches the single model (F1 \NSysFourF against \NSysOneF), while the graph agent and message passing are less accurate; a variant of the graph agent developed on a separate set did not close the gap, and both graph summaries and multi-turn tool use lowered accuracy. These results hold with reworded prompts and with a larger reasoning model, except that the larger model weakens message passing enough for the graph agent to beat it. Sound events improve F1 by \NDiffFourSound \NDiffFourSoundCI, in line with the annotators, who used sound or speech on most violent clips. Citations of stored observations are faithful, changing about half of the decisions when removed, but no more so with the graph than with flat text, and the agent pipeline costs about \NCostRatio times the GPU time of the single model. We also report two artefacts of vision-language describers that mislead pipelines built on them.
\end{abstract}

\section{Introduction}
\label{sec:intro}

Operators who review video footage for violence work under a budget: they cannot watch everything, so a system that ranks or flags clips is useful only if its decisions are accurate and can be checked. Large language models now make such systems possible without training. A video can be described in text and judged by a language model \citep{zanella2024lavad}, an agent can call tools and query a vision model \citep{yang2025panda,bekit2026qvad}, and several agents can share observations before one of them decides \citep{duan2026agenticvau}. In parallel, single audio-visual language models can now take a whole clip, picture and sound, and answer directly \citep{qwen3omni}.

Two questions are rarely asked of these designs. First, \emph{how} should the observations of several perception models reach the model that decides? Recent work proposes ever richer interfaces: shared boards \citep{salemi2025blackboard,han2025blackboard}, registries of evidence, multi-turn exploration with tools. These designs are usually compared with other systems that differ in many ways at once, so it is hard to tell what the interface itself contributes; and in other domains, multi-agent designs often fail to beat a single well-prompted model \citep{wang2024multiagentbounds,kim2025scalingagents}. Second, \emph{which evidence} matters? Language-model pipelines for violence and anomaly detection read frames or captions; sound and speech are at most named in a prompt \citep{bekit2026qvad}, although the benchmark they are evaluated on, XD-Violence \citep{wu2020xdviolence}, was built as an audio-visual dataset.

We address both questions with a controlled study on the XD-Violence test set. We fix one set of perception agents (a video describer, object detectors, a sound tagger, speech recognition and a speech tagger) and one reasoning model, and vary only how their observations reach the decision: as messages between agents, as a typed evidence graph explored with tools, or as flat text read in one pass. We compare these with a single audio-visual model given the raw clip. To know what the right answer and the right evidence are, we built a human reference: two annotators answered the operator's question on clips of the test set, marked the incidents in time, and recorded whether they relied on what they saw, on sound or on speech. Hypotheses and decision rules were fixed before any system was run, and when the graph agent underperformed we developed it further only on a separate development set drawn from the training videos.

The findings are mostly about what does \emph{not} help:
\begin{itemize}
  \item The single audio-visual model and flat text are equally accurate (F1 \NSysOneF and \NSysFourF; difference \NDiffFourMinusOne \NDiffFourMinusOneCI). The structured interfaces are less accurate: the graph agent minus flat text is \NDiffThreeMinusFour F1 \NDiffThreeMinusFourCI, and the graph agent does not beat message passing (\NDiffThreeMinusTwo \NDiffThreeMinusTwoCI). With paraphrased prompts the picture is the same; with a larger reasoning model flat text still leads, but the graph agent now beats message passing.
  \item On the development set, adding graph summaries to flat text lowers F1 from \NDevFourF to \NDevFourSummariesF, and turning the same information into a multi-turn exploration with tools lowers it to \NDevAttemptDF. Edges built from the agents' observations do not help, and in one variant they hurt.
  \item The graph agent does not fail for lack of information. On the violent clips it missed, it had usually been shown the evidence that flat text relied on (all of it in \NFailAllShown of \NFailViolent clips, some of it in \NFailPartShown). Its tools hand it fragments: it stops before opening the span that holds the incident, decides against evidence it has read, or, on normal clips, follows object detections taken out of the context that would discount them.
  \item Sound helps. On clips with sound, flat text with the sound events scores \NDiffFourSound F1 \NDiffFourSoundCI above flat text without them, consistent with the human reference, where sound or speech was used on \NAWithSoundSoundOrSpeech\% and \NBWithSoundSoundOrSpeech\% of the violent clips with sound.
  \item The systems that cite stored observations cite evidence their decisions depend on: hiding the cited observations changes \NFaithFourCited\% of flat-text decisions against \NFaithFourRandom\% for random ones. This holds as much for flat text as for the graph agent, and the single model offers nothing comparable to check.
  \item Counting the perception models, the agent pipeline costs about \NCostRatio times the GPU time of the single model; it saves only when stored observations are reused.
\end{itemize}

Our contributions are: (i) a human reference for XD-Violence that records the operator decision, the incident segments and the evidence modality used, with agreement between two annotators (Section~\ref{sec:human}); (ii) a controlled comparison of four ways for the same perception agents to reach a decision, with a development-set protocol, deletion tests of faithfulness and an analysis of why the graph agent errs (Sections~\ref{sec:method} and~\ref{sec:results}); and (iii) evidence that, on this task, flat text matches a strong single audio-visual model while structured agent interfaces cost accuracy, together with two artefacts of vision-language describers that can mislead any pipeline built on them (Section~\ref{sec:results-artefacts}).

\section{Related work}
\label{sec:related}

\paragraph{Violence and anomaly detection with language models.} XD-Violence \citep{wu2020xdviolence} pairs video with its soundtrack, and trained audio-visual models exploit both \citep{ghadiya2024crossmodal,wu2025avadclip}. Trained vision-language methods reach high frame-level AP on it, \NLitVadClipAP for VadCLIP \citep{wu2024vadclip} and \NLitHolmesVauAP for the instruction-tuned Holmes-VAU \citep{zhang2025holmesvau}. Training-free methods trade accuracy for not needing labelled data: LAVAD describes frames and lets a language model score them (\NLitLavadAP AP, \citealp{zanella2024lavad}); EventVAD builds a graph over frames before asking a multimodal model (\NLitEventVadAP, \citealp{shao2025eventvad}); QVAD lets an agent pose questions to a small vision model (\NLitQvadAP, \citealp{bekit2026qvad}); PANDA adds retrieval, tools and reflection to a single agent (\NLitPandaAP, \citealp{yang2025panda}); AgenticVAU has several agents share observations in a registry before deciding \citep{duan2026agenticvau}; VTO trains a tool-using agent with reinforcement learning and reports that such agents tend to stop exploring too early \citep{wang2026vto}, a failure we also observe without training (Section~\ref{sec:results-failures}). These pipelines read frames or captions; in the training-free ones sound is at most mentioned in the prompt, and none we know of feeds transcribed speech to a violence or anomaly decision. Transcripts are used for a neighbouring task: MARS gives a vision-language model sampled frames and a transcription to detect hateful video without training \citep{yang2026mars}. Our systems are training-free and give one score per clip, so they are not built to compete on frame-level AP; we report it for comparability (Section~\ref{sec:results-main}) and focus on the decision an operator needs.

\paragraph{How agents share information.} Blackboard architectures, in which agents post observations to a shared store read by others, have returned in language-model systems and were reported to beat agents directing each other \citep{salemi2025blackboard,han2025blackboard}. Other work questions whether several agents help at all: a single agent with strong prompts nearly matches multi-agent discussion \citep{wang2024multiagentbounds}, coordination helps on parallel tasks and hurts on sequential ones \citep{kim2025scalingagents}, and multi-agent systems fail in recurring ways that can be catalogued \citep{cemri2025mast}. These studies are text-only. To our knowledge, no study holds a set of multimodal perception agents fixed and varies only the interface through which their observations reach the decision, which is what Section~\ref{sec:results} does.

\paragraph{Human references for explanation.} Benchmarks for video anomaly detection label which frames are abnormal; they do not say which evidence a person would use to decide. Our human reference adds that, so that a system's reliance on sound or speech, and the segments it cites, can be compared with people's.

\section{Human reference}
\label{sec:human}

The XD-Violence test set \citep{wu2020xdviolence} labels each clip as normal or as showing one or more classes of violence. The decision our systems support is a different one: \emph{would a security operator need to look at this clip?} We call a clip for which the answer is yes \emph{review warranted}. Before using the violence labels as a stand-in, or proxy, for this decision, we asked human annotators to make it directly. The human reference serves three purposes. It measures how well the labels match the decision we care about; it records which kinds of evidence (what is seen, non-speech sound, speech) people rely on; and it gives time segments against which the explanations of the systems can be checked.

\subsection{Protocol}
\label{sec:human-protocol}

For each clip the annotator answers one question, ``Would a security operator need to look at this clip?'', with Yes, No or Unsure, and gives a confidence from 1 (guess) to 3 (certain). The protocol defines Yes as a clip that shows, or gives credible signs of, violence or danger to people: fighting, assault or abuse; weapons used or pointed at people; explosions or fire that threaten people; crowds attacking people or property; serious traffic or industrial accidents. Sound or speech alone can be enough, for example screams of fear, gunshots or explicit threats. No covers normal activity, crowds without aggression and sport played under its rules. Unsure is reserved for clips the annotator honestly cannot decide, such as ambiguous play fighting or a distant event that cannot be made out. Clips are judged as if they were real footage from a monitored place, even when they are clearly taken from a film or a news report.

For a Yes, the annotator marks every incident as a time segment and states, for each segment, the evidence they relied on: \emph{visual} (what is seen), \emph{sound} (non-speech audio such as screams, gunshots or crashes) or \emph{speech} (what people say). The question is what the annotator used, not what the clip contains. A category (fighting, shooting, riot, abuse, accident, explosion, other) and a short note are optional. For an Unsure, the annotator states what caused the doubt; segments are optional. A No needs nothing further.

Annotation was done in a local web application. Clips are shown with sound under random codes, because the XD-Violence file names contain the labels; the labels were not available to the annotators until all their answers were final. The application records the share of each clip that was played and the active time spent on it, saves every answer at once and lets the annotator return to an earlier clip. The protocol was fixed before annotation began, and each annotator's answers were frozen before they were compared with the labels or with the other annotator.

\subsection{Sample and annotators}
\label{sec:human-sample}

The test set has \NTestClips clips. One shooting clip was used to explain the task to the annotators and is excluded, which leaves \NPoolClips. Annotator~A annotated \NAClips clips drawn at random within each XD-Violence class (the first class listed for a clip), with at least \NAClassMin clips per class. To measure consistency, \NARepeats of these clips were shown a second time without any marker, at least \NAGapDesign positions after their first showing (the smallest gap in the session was \NAMinGap), for \NAItems items in total. Annotator~B annotated \NBItems clips, each shown once: \NBShared drawn at random within each class from annotator~A's sample, to measure agreement between the two annotators, and \NBNew that annotator~A had not seen, drawn within each class with at least \NBNewClassMin per class. Both kinds were mixed in random order under new codes, and annotator~B did not see annotator~A's answers. Table~\ref{tab:human-sample} gives the sample by class: \NDistinctClips distinct clips, \NBShared of them with two annotators.

Annotator~A is one of the authors and knew the design of the study; annotator~B took no part in the study and was new to the task. Annotator~A worked \NAHours hours (median \NAMedianMinutes minutes per item), annotator~B \NBHours hours (median \NBMedianMinutes minutes). All of annotator~A's items were played to at least \NAMinWatch\%; annotator~B played one clip to \NBMinWatch\% and answered Unsure on it, and every other clip to at least \NWatchThreshold\%.

\begin{table}[t]
  \centering
  \caption{Clips of the human reference by XD-Violence class (first class listed for the clip).}
  \label{tab:human-sample}
  \begin{tabular}{lrrrr}
\toprule
XD-Violence class & Annotator A & B, shared with A & B, new & Distinct clips \\
\midrule
Normal & 36 & 9 & 6 & 42 \\
Fighting & 15 & 4 & 3 & 18 \\
Shooting & 10 & 3 & 2 & 12 \\
Riot & 13 & 4 & 3 & 16 \\
Abuse & 3 & 3 & 2 & 5 \\
Car accident & 12 & 4 & 2 & 14 \\
Explosion & 11 & 3 & 2 & 13 \\
\midrule
Total & 100 & 30 & 20 & 120 \\
\bottomrule
\end{tabular}

\end{table}

\subsection{Audio content of the test set}
\label{sec:human-audio}

Every test clip has an audio track, but a track may hold only silence or music. To know on which clips audio evidence can matter at all, we measured the level of each second of audio and ran an audio event tagger, the Audio Spectrogram Transformer \citep{gong2021ast} trained on AudioSet \citep{gemmeke2017audioset}, on 10-second windows every 5 seconds. With thresholds fixed before the run, a clip is \emph{near-silent} if fewer than 10\% of its seconds are louder than $-50$~dBFS; otherwise it contains \emph{speech} if the speech classes score at least 0.5 in at least 10\% of its windows, it is \emph{music only} if music scores at least 0.5 in at least 80\% of its windows, and it contains \emph{other sound} in all remaining cases. We call clips with speech or other sound the \emph{with-sound subset}.

Table~\ref{tab:audio-content} shows that \NAudioWithSound of the \NPoolClips clips (\NAudioWithSoundShare\%) belong to the with-sound subset; \NAudioMusicOnly hold only music and \NAudioNearSilent is near-silent. Speech is present in most clips because most of the material comes from films. Sound is spread evenly across classes: \NAudioViolentSound\% of violent clips and \NAudioNormalSound\% of normal clips belong to the with-sound subset (Fisher's exact test, $p = \NAudioFisherP$). A system therefore cannot separate violent from normal clips by the mere presence of sound.

\begin{table}[t]
  \centering
  \caption{Audio content of the \NPoolClips test clips by XD-Violence class.}
  \label{tab:audio-content}
  \begin{tabular}{lrrrrr}
\toprule
XD-Violence class & Clips & Speech & Other sound & Music only & Near-silent \\
\midrule
Normal & 300 & 256 & 30 & 14 & 0 \\
Fighting & 120 & 97 & 15 & 8 & 0 \\
Shooting & 83 & 63 & 17 & 3 & 0 \\
Riot & 101 & 95 & 4 & 2 & 0 \\
Abuse & 8 & 8 & 0 & 0 & 0 \\
Car accident & 96 & 79 & 13 & 4 & 0 \\
Explosion & 91 & 64 & 24 & 2 & 1 \\
\midrule
All & 799 & 662 & 103 & 33 & 1 \\
\bottomrule
\end{tabular}

\end{table}

\subsection{Results}
\label{sec:human-results}

\paragraph{Are the labels a valid proxy?}
We compare Yes with a violent label and No with a normal label, on the first showing of each clip, and report Cohen's kappa \citep{cohen1960kappa} with a 95\% bootstrap interval over clips \citep{efron1993bootstrap}. On the \NADecided clips annotator~A decided (\NAUnsure answered Unsure), kappa is \NAKappa (\NAKappaLo to \NAKappaHi) and agreement is \NAAcc\%. Counting Unsure as Yes gives \NAKappaUnsureYes, counting it as No gives \NAKappaUnsureNo. Annotator~B agrees with the labels on all \NBDecided clips they decided and answered Unsure on \NBUnsure. Table~\ref{tab:proxy-class} shows that every violent class agrees fully for both annotators.

All differences go in the same direction. Annotator~A answered Yes on \NAYesNormal normal clips, a girl in a tree with men trying to pull her down and people on motorcycles closing in on a car to attack its occupants, and Unsure on two normal clips showing threatening stares and a tense confrontation. The one Unsure on a violent clip is an explosion: a dead whale bursting while it is cut open, with no danger to people. Review warranted is therefore slightly wider than violent, since it includes threats and possible escalation, and it leaves out an explosion that endangers no one. On this sample the violence labels are a valid proxy for the decision, which supports their use as the main benchmark.

\begin{table}[t]
  \centering
  \caption{Agreement with the XD-Violence labels by class. ``Agrees'' counts decided clips (Yes or No) whose answer matches the label; Unsure answers are listed separately.}
  \label{tab:proxy-class}
  \begin{tabular}{lrrrr}
\toprule
XD-Violence class & A agrees & A unsure & B agrees & B unsure \\
\midrule
Normal & 32/34 & 2 & 14/14 & 1 \\
Fighting & 15/15 & 0 & 7/7 & 0 \\
Shooting & 10/10 & 0 & 5/5 & 0 \\
Riot & 13/13 & 0 & 7/7 & 0 \\
Abuse & 3/3 & 0 & 5/5 & 0 \\
Car accident & 12/12 & 0 & 6/6 & 0 \\
Explosion & 10/10 & 1 & 4/4 & 1 \\
\bottomrule
\end{tabular}

\end{table}

\paragraph{Do two annotators agree?}
On the \NABShared shared clips, the two annotators gave the same answer on \NABSame (kappa over Yes, No and Unsure \NABKappa, 95\% interval \NABKappaLo to \NABKappaHi; \NABKappaBinary on the \NABBinaryN clips where neither answered Unsure). Both answered Unsure, independently, on the same \NABBothUnsure clips: the threatening stares and the exploding whale. The only disagreement is the clip with motorcycles around a car, which annotator~A found review warranted and annotator~B did not.

On the \NABBothYes clips both annotators called Yes, we compare where they placed the incidents. We take the union of each annotator's segments on the time axis and compute the intersection over union of the two. Its median is \NABIoUMedian (mean \NABIoUMean, lowest \NABIoUMin), and \NABIoUHalf of the \NABBothYes clips reach 0.5 or more. The evidence the two annotators relied on also matches: they agree on whether visual evidence was used on \NABVisualSame clips, on sound on \NABSoundSame and on speech on \NABSpeechSame, and they share at least one segment category on \NABCategoryShared.

\paragraph{Is an annotator consistent?}
Annotator~A gave the same answer to all \NARepeats repeated clips (kappa \NSelfKappa). This is an upper bound: the annotator recognised at least \NSelfRecognised repeats during the session.

\paragraph{Which evidence do people use?}
Annotator~A relied on sound or speech for \NAAllSoundOrSpeech\% of the \NAAllYesClips clips they found review warranted (sound \NAAllSound\%, speech \NAAllSpeech\%, visual evidence on all of them), annotator~B for \NBAllSoundOrSpeech\% of \NBAllYesClips (sound \NBAllSound\%, speech \NBAllSpeech\%). On the with-sound subset these shares are \NAWithSoundSoundOrSpeech\% of \NAWithSoundYesClips clips and \NBWithSoundSoundOrSpeech\% of \NBWithSoundYesClips, almost unchanged. The annotators marked \NASegments and \NBSegments segments on these clips, \NASegmentsPerYes and \NBSegmentsPerYes per clip.

\paragraph{Does an audio tagger hear what people heard?}
We checked the tagger of Section~\ref{sec:human-audio} against the segments of the clips found review warranted. A segment counts as heard if one of the violence-related event groups (gunshot, explosion, scream, crash, crowd, siren, impact) scores at least 0.3 in a window that overlaps it. Of the \NTagTicked segments where an annotator ticked sound, \NTagHeardTicked\% are heard, against \NTagHeardUnticked\% of the \NTagUnticked segments where sound was not ticked (\NTagHeardLowTicked\% and \NTagHeardLowUnticked\% at a threshold of 0.1). Table~\ref{tab:tagger} shows that this depends strongly on the kind of incident: the tagger hears most shootings and about half of the explosions, but few fights and very little abuse, whose sounds (blows, cries) are poorly covered by the AudioSet classes. Speech is detected in \NTagSpeechHeardTicked\% of the \NTagSpeechTicked segments where speech was ticked and in \NTagSpeechHeardUnticked\% of the others, so the presence of speech says little; what matters is what is said.

\begin{table}[t]
  \centering
  \caption{Segments where an annotator ticked sound, by the category the annotator gave, and the share in which the tagger finds a violence-related sound event at two score thresholds.}
  \label{tab:tagger}
  \begin{tabular}{lrrrl}
\toprule
Segment category & Segments & Event at 0.3 (\%) & Event at 0.1 (\%) & Most frequent event \\
\midrule
Shooting & 37 & 76 & 95 & gunshot \\
Explosion & 21 & 52 & 81 & explosion \\
Riot & 25 & 40 & 72 & crowd \\
Accident & 34 & 38 & 53 & crash \\
Fighting & 24 & 25 & 83 & explosion \\
Abuse & 18 & 11 & 39 & crash \\
\bottomrule
\end{tabular}

\end{table}

\paragraph{Consequences for the evaluation.}
Because the labels are a valid proxy on this sample, the systems are evaluated against them on the full test set, and the human reference is used for what the labels do not measure: where the evidence is and which modality carries it. Questions about the value of audio are answered on the with-sound subset and reported per class, since off-the-shelf audio events are informative for shootings and explosions but not for fights and abuse. Speech enters the systems through its content (transcription and tagging of threats or distress), not through its presence.

\subsection{Limitations of the human reference}
\label{sec:human-limits}

The reference rests on two annotators, one of whom is an author who knew the design of the study; the other was new to the task. Most sampled clips are clear-cut, so near-perfect agreement with the labels says little about borderline cases, and the abuse class has only \NDistinctAbuse distinct clips, too few for per-class claims. Annotator~B saw no repeated clips, so consistency is measured for annotator~A only, and that measure is an upper bound. The clips come mostly from films and online videos, not from surveillance cameras. The audio categories depend on one tagger and on thresholds that were fixed in advance but not tuned.

\section{Method}
\label{sec:method}

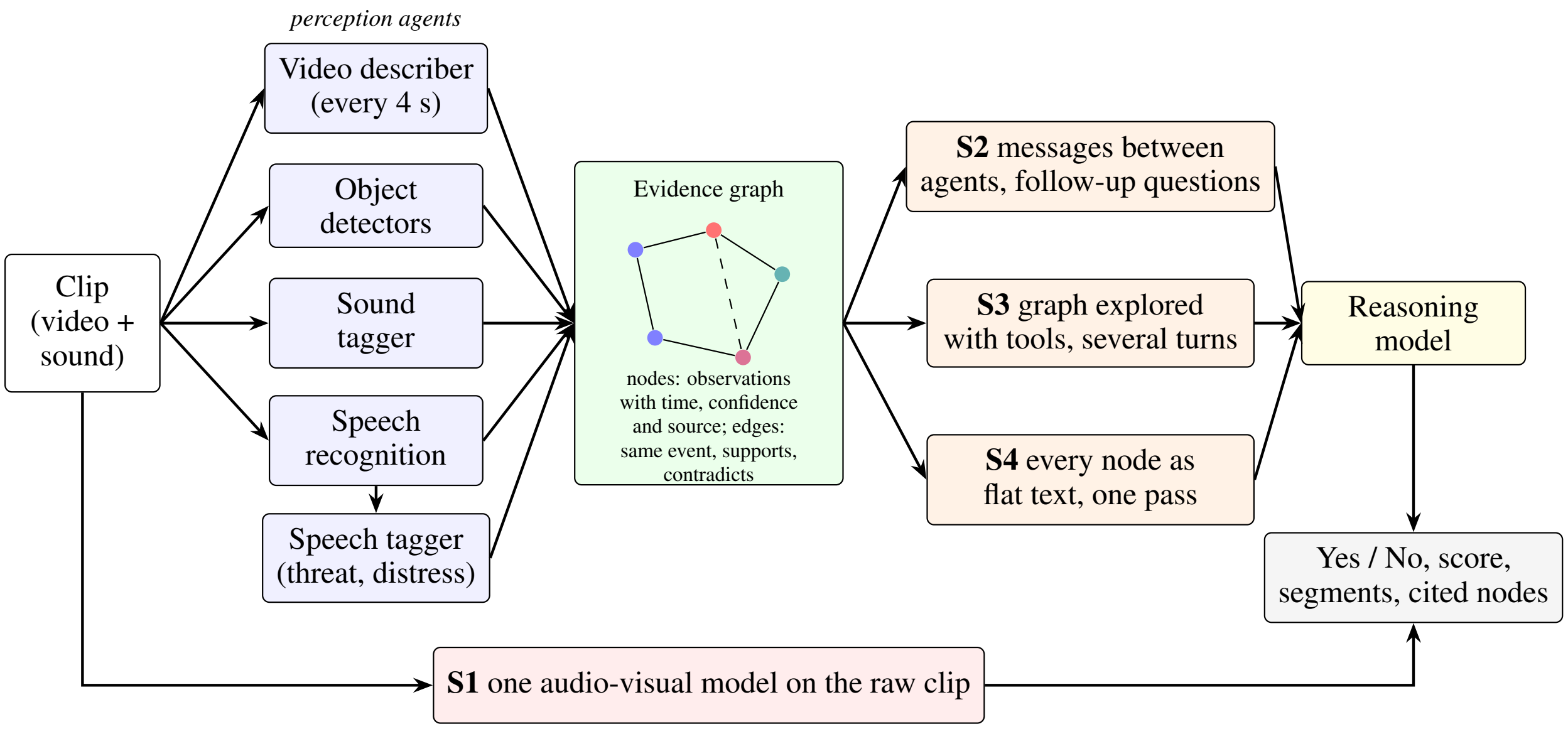
\begin{figure*}[t]
  \centering
  \resizebox{\textwidth}{!}{\begin{tikzpicture}[
  font=\small,
  box/.style={draw, rounded corners=2pt, align=center, inner sep=4pt, minimum height=2.2em},
  agent/.style={box, fill=blue!6, minimum width=6.2em},
  sys/.style={box, fill=orange!10, minimum width=9.5em},
  store/.style={box, fill=green!8, minimum width=7.8em, minimum height=9.4em},
  result/.style={box, fill=gray!8, minimum width=8.5em},
  arr/.style={-{Stealth[length=2mm]}, thick},
  lab/.style={font=\scriptsize, fill=white, inner sep=1pt}
]
  \node[box, minimum width=4.5em, minimum height=4em] (clip) at (0,0) {Clip\\(video +\\sound)};

  \node[agent] (vid) at (3.0, 2.4) {Video describer\\(every 4 s)};
  \node[agent] (det) at (3.0, 1.2) {Object\\detectors};
  \node[agent] (snd) at (3.0, 0.0) {Sound\\tagger};
  \node[agent] (asr) at (3.0,-1.2) {Speech\\recognition};
  \node[agent] (tag) at (3.0,-2.4) {Speech tagger\\(threat, distress)};
  \foreach \a in {vid,det,snd,asr} \draw[arr] (clip.east) -- (\a.west);
  \draw[arr] (asr.south) -- (tag.north);
  \node[font=\scriptsize\itshape, above=0pt of vid] {perception agents};

  \node[store] (graph) at (6.4,0) {};
  \node[font=\scriptsize, align=center] at (6.4,1.35) {Evidence graph};
  \begin{scope}[shift={(6.4,0.2)}, every node/.style={circle, inner sep=1.6pt}]
    \node[fill=blue!50]  (n1) at (-0.75, 0.55) {};
    \node[fill=red!55]   (n2) at ( 0.05, 0.75) {};
    \node[fill=teal!60]  (n3) at ( 0.75, 0.30) {};
    \node[fill=blue!50]  (n4) at (-0.55,-0.35) {};
    \node[fill=purple!55](n5) at ( 0.35,-0.55) {};
    \draw (n1)--(n2) (n2)--(n3) (n1)--(n4) (n4)--(n5) (n3)--(n5);
    \draw[dashed] (n2)--(n5);
  \end{scope}
  \node[font=\tiny, align=center] at (6.4,-1.05) {nodes: observations\\with time, confidence\\and source; edges:\\same event, supports,\\contradicts};
  \foreach \a in {vid,det,snd,asr,tag} \draw[arr] (\a.east) -- (graph.west);

  \node[sys] (s2) at (10.3, 1.6) {\textbf{S2} messages between\\agents, follow-up questions};
  \node[sys] (s3) at (10.3, 0.0) {\textbf{S3} graph explored\\with tools, several turns};
  \node[sys] (s4) at (10.3,-1.6) {\textbf{S4} every node as\\flat text, one pass};
  \foreach \s in {s2,s3,s4} \draw[arr] (graph.east) -- (\s.west);

  \node[box, fill=yellow!12, minimum width=6.5em] (llm) at (13.6,0) {Reasoning\\model};
  \foreach \s in {s2,s3,s4} \draw[arr] (\s.east) -- (llm.west);
  \node[result] (ans) at (13.6,-2.6) {Yes / No, score,\\segments, cited nodes};
  \draw[arr] (llm.south) -- (ans.north);

  \node[sys, fill=red!7, minimum width=13em] (s1) at (6.4,-3.7) {\textbf{S1} one audio-visual model on the raw clip};
  \draw[arr] (clip.south) |- (s1.west);
  \draw[arr] (s1.east) -| (ans.south);
\end{tikzpicture}}
  \caption{The compared systems. Perception agents write their observations into one evidence graph per clip. S2, S3 and S4 pass the same observations to the same reasoning model in three ways; S1 gives the raw clip to one audio-visual model. All four answer the same question and return a decision, a score and time segments; S3 and S4 also cite the nodes their answer rests on.}
  \label{fig:overview}
\end{figure*}

Figure~\ref{fig:overview} shows the systems. All of them answer the same question about each clip, the one the annotators answered: does the clip need a security operator's attention because it shows violence or danger? The prompt names the six XD-Violence classes, asks the model to judge what is shown even in films and news footage, and asks for Yes or No on the first line and the time segments of the incident on the second. The \emph{score} of a system on a clip is the probability of Yes divided by the probability of Yes plus No on the first answer token; the \emph{decision} is the first word of the answer. No model is trained.

\subsection{Evidence graph}
\label{sec:method-graph}

The perception agents described below do not talk to each other. Each writes what it observes into an \emph{evidence graph}, one per clip. A node is one observation, never an entity: a description of four seconds of video, an object detection, a sound event, a line of transcribed speech. Each node carries a time span in the clip, a confidence, and its provenance: the agent, the model and its version, and the position in the input. Edges are typed relations between observations from different agents whose time spans overlap: \emph{same event} (both nodes assert a violence-related or scene concept, graded by the overlap of their spans), \emph{supports} (their concepts agree, graded by the lower confidence), \emph{contradicts} (a video description denies a violence-related concept that the other node asserts, for example ``no weapons are visible'' beside a gunshot), and \emph{derived from} (a speech tag and the transcript line it was read from). Concepts are assigned by a fixed word list per concept (gun, knife, fire, fight, crash, injury, distress, threat, emergency, crowd, vehicle), applied sentence by sentence, with a sentence containing a negation denying its concepts instead of asserting them. Graphs are stored as append-only files, one per agent and clip, under a neutral clip identifier, so that nothing a system reads reveals the file name or label.

\subsection{Perception agents}
\label{sec:method-agents}

\begin{itemize}
  \item \textbf{Video.} One frame per second. An object detector for people and weapons (YOLOv8-L trained in earlier work of the authors on another domain) and an open-vocabulary detector \citep{cheng2024yoloworld} for fire, smoke, vehicles and police; detections of the same object in nearby seconds form one node. A vision-language model \citep{qwen3vl} describes each four seconds of video in at most 40 words.
  \item \textbf{Sound.} An AudioSet tagger \citep{gong2021ast} on 10-second windows every 5 seconds; consecutive windows in which a violence-related event (gunshot, explosion, scream, crash, crowd, siren, impact) scores at least 0.1 form one node, with the peak score as confidence.
  \item \textbf{Speech.} Whisper large-v3 \citep{radford2023whisper} transcribes speech in any language, one node per segment. The reasoning model then reads each line with its neighbours and labels it as a threat of violence, distress, a report of violence, or none; the label probabilities come from the next-token probabilities, so each tag carries a graded confidence.
\end{itemize}

The graphs of the \NSysClips test clips hold \NGraphNodes nodes (median \NGraphNodesMedian per clip): \NGraphNCaption video descriptions, \NGraphNDetection detections, \NGraphNSoundEvent sound events, \NGraphNTranscriptSegment transcript lines and \NGraphNSpeechTag speech tags; and \NGraphESameEvent same-event, \NGraphESupports supports, \NGraphEContradicts contradicts and \NGraphEDerivedFrom derived-from edges.

\subsection{Compared systems}
\label{sec:method-systems}

\begin{description}
  \item[S1, one omni model.] An audio-visual language model \citep{qwen3omni} receives the frames at one per second (at most 240, evenly spaced on longer clips) and the soundtrack, and answers directly. No agents, no graph. Because it often gave times past the end of the clip, we also run \textbf{S1-ts}, in which the clip is given in 20-second chunks, each preceded by a text label with its start and end time, so that the model sees absolute time as the text-based systems do.
  \item[S2, message passing.] The agents are grouped into three (video, sound, speech), and each group is played by the reasoning model given only that group's nodes as text. Each sends a short report; the reasoning agent may ask each agent follow-up questions before answering. No graph and no shared node identifiers.
  \item[S3, graph agent.] The reasoning agent reads the graph through tools: nodes in a time window, nodes asserting a concept, the edges of a node, the provenance of a node. It may call up to eight tools before answering, and cites the node identifiers its answer rests on. \textbf{S3 v2} was developed afterwards on a separate development set (Section~\ref{sec:results-dev}): it starts from all video descriptions, a timeline of the concepts each agent asserts per 10 seconds, and the groups of nodes from different agents joined by supports edges, and may call up to twelve tools.
  \item[S4, flat text.] Every node of every agent, one line each, sorted by time, in a single prompt; one pass, no tools. It cites node identifiers like S3.
\end{description}

S2, S3 and S4 use the same reasoning model, Gemma~4 12B \citep{gemma4} in bfloat16, with greedy decoding; Section~\ref{sec:results-robust} repeats them with a larger model. All four systems therefore differ only in how the same evidence reaches the decision: as raw audio and video (S1), as messages (S2), as a graph explored with tools (S3), or as flat text (S4).

\subsection{Evaluation}
\label{sec:method-eval}

The evaluation set is the whole test set, \NSysClips clips (\NSysViolent violent, \NSysNormal normal), fixed with a hash of the file list before any system was run. We report, against the XD-Violence labels: the area under the ROC curve (AUC) of the score; F1 of the decision; frame-level average precision (AP) over every frame of every clip at 24 frames per second, where a frame receives the clip score if it lies in a segment the system named (or if the system named none) and 0 otherwise; and \emph{grounding precision and recall}, the share of predicted seconds that fall inside the labelled violent frames and the share of labelled violent seconds that are covered, on violent clips the system called Yes. Segments are cut at the end of the clip for every system. We also report F1 against annotator~A's decisions on the clips of Section~\ref{sec:human}. Differences between systems are paired and come with 95\% bootstrap confidence intervals over clips. \emph{Faithfulness} is measured by deletion: we hide the nodes a system cited and run it again, and compare how often its decision changes with hiding as many randomly chosen uncited nodes. Hypotheses and decision rules were fixed before any system was run.

\section{Results}
\label{sec:results}

\subsection{Main comparison}
\label{sec:results-main}

Table~\ref{tab:systems} compares the systems on the \NSysClips test clips. The single omni model is the most accurate (AUC \NSysOneAUC, F1 \NSysOneF). Flat text (S4) is as accurate: its F1 differs from S1's by \NDiffFourMinusOne \NDiffFourMinusOneCI, and from S1-ts's by \NDiffFourMinusOneTs \NDiffFourMinusOneTsCI. Both structured designs are less accurate. The graph agent S3 is below flat text by \NDiffThreeMinusFour F1 \NDiffThreeMinusFourCI, and it does not beat message passing: S3 minus S2 is \NDiffThreeMinusTwo \NDiffThreeMinusTwoCI, so the hypothesis that the shared graph improves on message passing is not supported with this reasoning model (with a larger one it is; Section~\ref{sec:results-robust}).

\begin{table*}[t]
  \centering
  \caption{Systems on the \NSysClips test clips. Frame AP: every frame at 24 frames per second. Grounding: share of predicted seconds inside labelled violent frames (precision) and share of labelled violent seconds covered (recall), on violent clips called Yes. Tokens: reasoning tokens per clip (prompt and answer); perception cost is in Table~\ref{tab:cost}. F1 vs A: against annotator~A's decisions on the clips of Section~\ref{sec:human}.}
  \label{tab:systems}
  \resizebox{\textwidth}{!}{\begin{tabular}{lccccccc}
\toprule
System & AUC [95\% CI] & F1 [95\% CI] & Frame AP & Ground.\ prec. & Ground.\ rec. & Tokens & F1 vs A \\
\midrule
S1 omni & 0.997 [0.994, 0.999] & 0.968 [0.956, 0.978] & 0.523 & 0.645 & 0.560 & 9{,}395 & 0.954 \\
S1-ts omni, time labels & 0.997 [0.994, 0.999] & 0.966 [0.954, 0.977] & 0.573 & 0.551 & 0.803 & 9{,}097 & 0.985 \\
S2 messages & 0.941 [0.923, 0.957] & 0.930 [0.913, 0.946] & 0.490 & 0.637 & 0.717 & 8{,}747 & 0.913 \\
S3 graph, v1 & 0.951 [0.934, 0.966] & 0.935 [0.917, 0.950] & 0.443 & 0.652 & 0.678 & 6{,}634 & 0.952 \\
S3 graph, v2 & 0.960 [0.946, 0.972] & 0.924 [0.906, 0.941] & 0.543 & 0.658 & 0.762 & 9{,}574 & 0.936 \\
S4 flat text & 0.992 [0.987, 0.996] & 0.967 [0.956, 0.978] & 0.538 & 0.627 & 0.774 & 3{,}255 & 0.961 \\
\bottomrule
\end{tabular}
}
\end{table*}

Grounding does not favour the graph. With absolute time given in the input, the omni model covers more of the violent time than any other system (S1-ts recall \NSysOneTsGroundR against \NSysFourGroundR for S4) and has the highest frame-level AP (\NSysOneTsFrameAP). The text-based systems name tighter segments (precision \NSysThreeBGroundP for S3 v2, \NSysFourGroundP for S4, against \NSysOneTsGroundP for S1-ts). On frame-level AP all systems stay below published training-free results on this benchmark \citep{zanella2024lavad}, because each gives one score per clip rather than a score per frame; naming segments changes frame AP little (S4: \NSysFourFrameAP with segments, \NSysFourFrameAPNoSeg without).

Against annotator~A's decisions the ranking is the same: S1-ts reaches F1 \NSysOneTsFvsA, S4 \NSysFourFvsA, S3 \NSysThreeFvsA and S2 \NSysTwoFvsA.

\subsection{What helps and what does not}
\label{sec:results-ablation}

Table~\ref{tab:pairs} lists the paired differences, each written as the first system minus the second. Sound helps. On clips with sound, S4 with sound events scores \NDiffFourSound F1 \NDiffFourSoundCI above S4 without them, and \NDiffFourSoundSpeech \NDiffFourSoundSpeechCI above S4 without both sound events and speech; speech alone adds \NDiffFourSpeech \NDiffFourSpeechCI, which is not significant. This agrees with the human reference, where sound or speech was used on \NAWithSoundSoundOrSpeech\% (annotator~A) and \NBWithSoundSoundOrSpeech\% (annotator~B) of the violent clips with sound. For S3 v1 sound events add \NDiffThreeSound \NDiffThreeSoundCI; for S3 v2 they add nothing measurable (\NDiffThreeBSound \NDiffThreeBSoundCI).

Edges do not help. S3 v1 with edges scores \NDiffThreeEdges \NDiffThreeEdgesCI above S3 v1 without them; S3 v1 asked for the edges of a node in only \NToolLinksThree of its \NToolCallsThree tool calls. For S3 v2, which builds its timeline and events from the edges, they \emph{lower} F1: with minus without is \NDiffThreeBEdges \NDiffThreeBEdgesCI.

\begin{table}[t]
  \centering
  \caption{Paired differences in F1 and AUC, with 95\% bootstrap confidence intervals over clips. ``With sound'': the clips whose soundtrack holds speech or other sound (Section~\ref{sec:human-audio}).}
  \label{tab:pairs}
  \small
  \begin{tabular}{llrcr}
\toprule
Comparison & Clips & $\Delta$F1 & 95\% CI & $\Delta$AUC \\
\midrule
S3 v1 $-$ S2 & all & +0.005 & [$-$0.011, +0.020] & +0.010 \\
S3 v2 $-$ S2 & all & $-$0.006 & [$-$0.021, +0.009] & +0.019 \\
S3 v1 $-$ S4 & all & $-$0.032 & [$-$0.046, $-$0.019] & $-$0.041 \\
S4 $-$ S1 & all & $-$0.001 & [$-$0.014, +0.012] & $-$0.005 \\
S4 $-$ S1-ts & all & +0.001 & [$-$0.012, +0.014] & $-$0.005 \\
S3 v2 $-$ S3 v1 & all & $-$0.011 & [$-$0.027, +0.005] & +0.009 \\
S3 v1 with $-$ without edges & all & +0.005 & [0.000, +0.011] & +0.004 \\
S3 v2 with $-$ without edges & all & $-$0.013 & [$-$0.024, $-$0.003] & +0.002 \\
S3 v1 with $-$ without sound events & with sound & +0.013 & [+0.003, +0.024] & +0.010 \\
S3 v2 with $-$ without sound events & with sound & $-$0.002 & [$-$0.012, +0.008] & +0.009 \\
S4 with $-$ without sound events & with sound & +0.012 & [+0.004, +0.021] & +0.005 \\
S4 with $-$ without speech & with sound & +0.005 & [$-$0.001, +0.011] & +0.001 \\
S4 with $-$ without sound and speech & with sound & +0.017 & [+0.007, +0.027] & +0.009 \\
S3 v1, Qwen3.6 $-$ Gemma & all & +0.004 & [$-$0.011, +0.020] & +0.028 \\
\bottomrule
\end{tabular}

\end{table}

\subsection{Robustness to the prompt and the reasoning model}
\label{sec:results-robust}

Two checks, added after the main results and not part of the hypotheses fixed in advance, ask whether the comparison depends on the wording of the prompts or on the reasoning model (Table~\ref{tab:robust}). First, we reran S3 v1 and S4 with paraphrased instructions: a reworded system prompt for S4 and a reworded opening paragraph for S3 (Appendix~\ref{app:prompts}), with the task question unchanged. Neither system moves: paraphrased minus original is \NRobParaFour F1 \NRobParaFourCI for S4 and \NRobParaThree \NRobParaThreeCI for S3, and with the paraphrases S3 is still below S4 by \NRobParaThreeMinusFour \NRobParaThreeMinusFourCI.

Second, we reran S2, S3 v1 and S4 with a larger reasoning model, Qwen3.6-35B-A3B \citep{qwen36}, with its thinking mode off. Flat text is again the most accurate: S3 minus S4 is \NRobQwenThreeMinusFour \NRobQwenThreeMinusFourCI, and S4 scores the same with either model (\NRobQwenFourMinusGemma \NRobQwenFourMinusGemmaCI). The graph agent explores more with the larger model (\NSysQwenCalls tool calls per clip against \NSysThreeCalls) and ranks clips better (AUC \NRobThreeQwenAUC), but its F1 is no higher (\NDiffQwenMinusThree \NDiffQwenMinusThreeCI) and it reads \NSysQwenTokens tokens per clip. One result does change: message passing loses accuracy with the larger model (Qwen3.6 minus Gemma is \NRobQwenTwoMinusGemma F1 \NRobQwenTwoMinusGemmaCI), so that with Qwen3.6 the graph agent \emph{does} beat message passing (\NRobQwenThreeMinusTwo \NRobQwenThreeMinusTwoCI). Whether the graph improves on message passing therefore depends on the reasoning model, while the advantage of flat text over both does not.

\begin{table}[t]
  \centering
  \caption{Robustness checks on the \NSysClips test clips. Paraphrased: the same model with reworded instructions (S2 not rerun). Qwen3.6: the same prompts with a larger reasoning model; \NRobTwoQwenUnparsed S2 and \NRobThreeQwenUnparsed S3 answers could not be parsed and count as errors.}
  \label{tab:robust}
  \small
  \begin{tabular}{lcccccc}
\toprule
 & \multicolumn{2}{c}{Gemma 4 12B} & \multicolumn{2}{c}{Gemma, paraphrased} & \multicolumn{2}{c}{Qwen3.6-35B-A3B} \\
\cmidrule(lr){2-3} \cmidrule(lr){4-5} \cmidrule(lr){6-7}
System & AUC & F1 & AUC & F1 & AUC & F1 \\
\midrule
S2 messages & 0.941 & 0.930 & -- & -- & 0.937 & 0.891 \\
S3 graph, v1 & 0.951 & 0.935 & 0.946 & 0.936 & 0.979 & 0.939 \\
S4 flat text & 0.992 & 0.967 & 0.988 & 0.970 & 0.988 & 0.962 \\
\bottomrule
\end{tabular}

\end{table}

\subsection{Developing S3 on a separate set}
\label{sec:results-dev}

Because S3 lost to flat text, we tried to improve it without touching the test clips. We drew a development set of \NDevClips clips from the XD-Violence training videos (\NDevNormal normal, \NDevViolent violent; half films and half web videos, like the test set; no file in common with it) and ran the same perception agents on it. One clip longer than any test clip was left out, leaving \NDevUsed. Before starting we fixed a rule: at most four variants, and the one with the best development F1 is run once on the test set. Table~\ref{tab:dev} shows the attempts. The selected variant reached \NDevAttemptCF on the development set (a second run gave \NDevReproF) against \NDevThreeOneF for S3 v1, but \NDevFourF for flat text. On the test set S3 v2 did not improve on S3 v1 (\NDiffThreeBMinusThree \NDiffThreeBMinusThreeCI).

The last two rows of Table~\ref{tab:dev} locate the loss. Adding the timeline and the event groups to flat text, still in one pass, lowers F1 from \NDevFourF to \NDevFourSummariesF; turning the same information into a multi-turn exploration with tools lowers it further, to \NDevAttemptDF. Both the graph summaries and the tool loop cost accuracy against reading everything at once.

\begin{table}[t]
  \centering
  \caption{Variants of S3 on the development set (\NDevUsed clips), in the order they were tried. The selected variant is the one run on the test set as S3 v2. The S4 rows are references; the one-pass row is a diagnostic and was not eligible for selection.}
  \label{tab:dev}
  \small
  \begin{tabular}{lrrrr}
\toprule
Variant & F1 & Precision & Recall & Tokens \\
\midrule
S3 v1 (as run on the test set) & 0.893 & 0.929 & 0.859 & 6{,}544 \\
a: whole-clip timeline and event tool & 0.880 & 0.849 & 0.913 & 6{,}831 \\
b: + corroborated detector cues, reading enforced & 0.888 & 0.919 & 0.859 & 11{,}010 \\
c: + all video descriptions up front (selected) & 0.907 & 0.912 & 0.902 & 9{,}603 \\
d: + every item up front & 0.896 & 0.901 & 0.891 & 11{,}946 \\
S4 + timeline and events, one pass (diagnostic) & 0.926 & 0.898 & 0.957 & 3{,}533 \\
S4 flat text & 0.946 & 0.936 & 0.957 & 3{,}152 \\
\bottomrule
\end{tabular}

\end{table}

\subsection{Why the graph agent errs}
\label{sec:results-failures}

To see why the tool loop costs accuracy, we examined every test clip that S3 v1 got wrong and flat text got right: \NFailClips clips, \NFailViolent of them violent clips that S3 missed and \NFailNormal normal clips that it flagged. Because the tools are deterministic, we first replayed each trace offline to recover exactly which nodes S3 was shown. Over all clips S3 saw about \NFailShownShareAll\% of a clip's nodes, in \NFailToolCallsAll tool calls on average. On the \NFailViolent missed violent clips, S3 had been shown every node that flat text cited in \NFailAllShown, some of them in \NFailPartShown, and none in only \NFailNeverShown. Missing evidence is therefore not the main cause.

We then read the \NFailClips traces and assigned each to one category (Table~\ref{tab:failures}); the categories were fixed before reading, except ``only faint evidence'', which we added while reading. Three causes account for most errors. S3 stopped before opening the time span that held the incident (\NFailEarlyStop clips). It read the decisive evidence and still answered No (\NFailMisjudged); Table~\ref{tab:example} shows one such clip, where every other system answered Yes. And on normal clips it was led by false object detections (\NFailDetectorLure of the \NFailNormal false alarms): a search for a concept returns the matching detections without the descriptions around them, whereas flat text shows each detection next to descriptions that make it implausible. In \NFailLabelDebatable clips the label itself is debatable under the annotation protocol, for example a film scene in which a man points a gun at a guard, which the protocol counts as Yes but the dataset labels normal. No error came from searching with words the lexicon does not cover. The tool interface thus hands the model fragments: too little of the clip in some cases, evidence out of context in others.

\begin{table}[t]
  \centering
  \caption{Errors of S3 v1 on test clips that flat text (S4) got right, one category per clip.}
  \label{tab:failures}
  \small
  \begin{tabular}{lrp{0.5\linewidth}}
\toprule
Category & Clips & Example \\
\midrule
Stopped before the evidence & 10 & stopped before the collision and the fight in the van \\
Read the evidence, decided wrongly & 9 & read a hockey fight with a stick strike, answered No \\
Led by false detections & 8 & Yes from RPG and pistol detections only \\
Label debatable & 9 & small fires and battery sparks, no people at risk \\
Only faint evidence & 4 & only cue a faint boom (0.14); descriptions show nothing \\
Searched with the wrong words & 0 & -- \\
\bottomrule
\end{tabular}

\end{table}

\begin{table*}[t]
  \centering
  \caption{One test clip (\NExDuration seconds, from a film, labelled shooting). Top: the \NExItems nodes of its graph, as S4 reads them. Middle: the same nodes on a timeline, one lane per kind of observation, with the edges the rules added (numbers as in the list above). Bottom: each system's answer. S3 opened the descriptions of the whole clip and searched for weapons, then answered No; all other systems answered Yes.}
  \label{tab:example}
  \begin{tabular}{p{0.95\linewidth}}
\toprule
\textbf{Evidence in the graph} \\
\texttt{\scriptsize [n1] 0-4s video description: A figure with a bright light emerges from a doorway. A man with bloodied face turns, then lunges forward. He stands in a dark, arched corridor, smoke billowing from his body.} \\
\texttt{\scriptsize [n2] 0-6s video detector saw 'person' (up to 1 at once, 5 s, conf 0.96)} \\
\texttt{\scriptsize [n3] 0-15s sound: gunshot (Machine gun, conf 0.28)} \\
\texttt{\scriptsize [n4] 1-2s speech (en, conf 0.40): "Oh, my God!"} \\
\texttt{\scriptsize [n5] 1-2s speech tagged distress, conf 1.00: "Oh, my God!"} \\
\texttt{\scriptsize [n6] 2-16s speech (en, conf 0.40): "Jerks. They should have shot me in the head, and enough times to make sure."} \\
\texttt{\scriptsize [n7] 2-16s speech tagged report, conf 1.00: "Jerks. They should have shot me in the head, and enough times to make sure."} \\
\texttt{\scriptsize [n8] 3-5s video detector saw 'pistol\_hand\_gun' (up to 1 at once, 2 s, conf 0.64)} \\
\texttt{\scriptsize [n9] 4-8s video description: A bloodied, disfigured figure stands in a dark, stone-walled room. The figure's head tilts back, mouth open as if screaming. The scene is dimly lit, with smoke or steam visible in the background.} \\
\texttt{\scriptsize [n10] 8-12s video description: Four surgeons in blood-stained gowns and caps look down from above at surgical lights. Their heads are visible, suggesting they are focused on a procedure below.} \\
\texttt{\scriptsize [n11] 11-17s video detector saw 'person' (up to 2 at once, 6 s, conf 0.92)} \\
\texttt{\scriptsize [n12] 12-16s video description: Surgeons in blood-splattered scrubs and gloves perform surgery. One surgeon holds a scalpel, another a surgical instrument. Blood drips from their hands and clothing. The scene is intense and graphic.} \\
\texttt{\scriptsize [n13] 16-16s speech (en, conf 0.40): "It..."} \\
\texttt{\scriptsize [n14] 16-17s video description: Surgical team in white gowns and masks, blood-splattered, hunched over patient. Bright overhead lights illuminate scene. One surgeon holds instrument near patient's body. Blood drips down gowns.} \\
\midrule
\textbf{The same nodes on a timeline}, with the edges added by rules \\
\makebox[\linewidth]
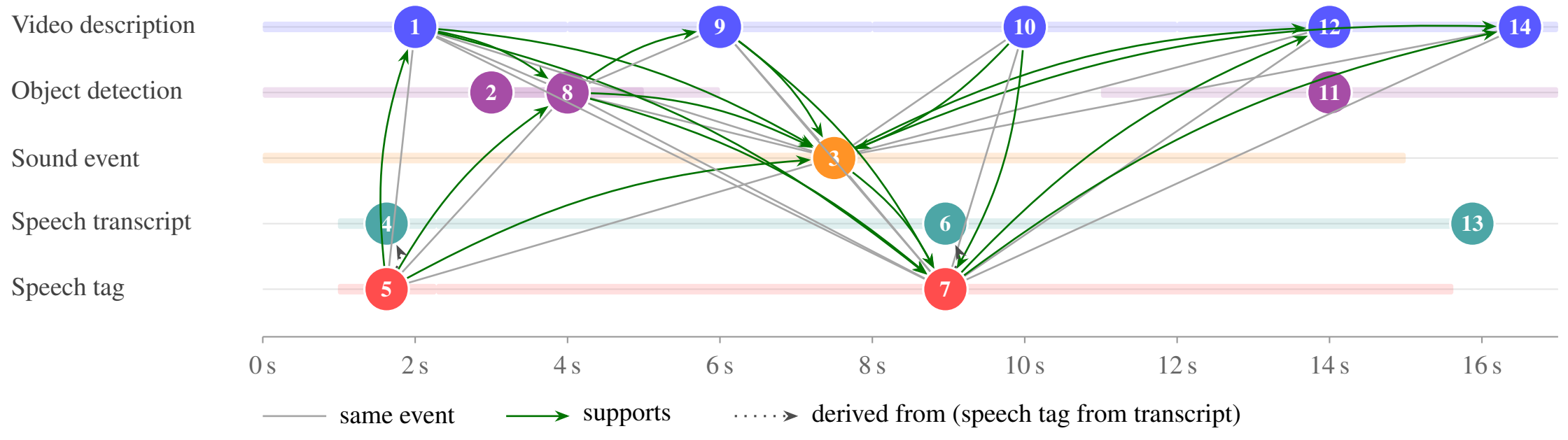
{\begin{tikzpicture}[font=\scriptsize,
  ev/.style={circle, inner sep=0pt, minimum size=4.2mm, draw=white, line width=0.6pt, text=white, font=\tiny\bfseries},
  same/.style={draw=black!35, line width=0.5pt},
  sup/.style={draw=green!45!black, line width=0.5pt, -{Stealth[length=1.4mm]}, bend left=12},
  der/.style={draw=black!70, line width=0.6pt, dotted, -{Stealth[length=1.4mm]}, bend right=25}]
\node[anchor=west, text=black!75] at (0,0.00) {Video description};
\draw[black!10] (2.50,0.00) -- (14.74,0.00);
\node[anchor=west, text=black!75] at (0,-0.62) {Object detection};
\draw[black!10] (2.50,-0.62) -- (14.74,-0.62);
\node[anchor=west, text=black!75] at (0,-1.24) {Sound event};
\draw[black!10] (2.50,-1.24) -- (14.74,-1.24);
\node[anchor=west, text=black!75] at (0,-1.86) {Speech transcript};
\draw[black!10] (2.50,-1.86) -- (14.74,-1.86);
\node[anchor=west, text=black!75] at (0,-2.48) {Speech tag};
\draw[black!10] (2.50,-2.48) -- (14.74,-2.48);
\fill[blue!65, opacity=0.18, rounded corners=0.6pt] (2.50,-0.05) rectangle (5.38,0.05);
\fill[violet!70, opacity=0.18, rounded corners=0.6pt] (2.50,-0.67) rectangle (6.82,-0.57);
\fill[orange!85, opacity=0.18, rounded corners=0.6pt] (2.50,-1.29) rectangle (13.30,-1.19);
\fill[teal!70, opacity=0.18, rounded corners=0.6pt] (3.21,-1.91) rectangle (4.14,-1.81);
\fill[red!70, opacity=0.18, rounded corners=0.6pt] (3.21,-2.53) rectangle (4.14,-2.43);
\fill[teal!70, opacity=0.18, rounded corners=0.6pt] (4.14,-1.91) rectangle (13.75,-1.81);
\fill[red!70, opacity=0.18, rounded corners=0.6pt] (4.14,-2.53) rectangle (13.75,-2.43);
\fill[violet!70, opacity=0.18, rounded corners=0.6pt] (4.66,-0.67) rectangle (6.10,-0.57);
\fill[blue!65, opacity=0.18, rounded corners=0.6pt] (5.38,-0.05) rectangle (8.26,0.05);
\fill[blue!65, opacity=0.18, rounded corners=0.6pt] (8.26,-0.05) rectangle (11.14,0.05);
\fill[violet!70, opacity=0.18, rounded corners=0.6pt] (10.42,-0.67) rectangle (14.74,-0.57);
\fill[blue!65, opacity=0.18, rounded corners=0.6pt] (11.14,-0.05) rectangle (14.02,0.05);
\fill[teal!70, opacity=0.18, rounded corners=0.6pt] (13.75,-1.91) rectangle (14.11,-1.81);
\fill[blue!65, opacity=0.18, rounded corners=0.6pt] (14.02,-0.05) rectangle (14.74,0.05);
\draw[black!40] (2.50,-2.93) -- (14.74,-2.93);
\draw[black!40] (2.50,-2.93) -- ++(0,-0.06) node[below, text=black!60] {0\,s};
\draw[black!40] (3.94,-2.93) -- ++(0,-0.06) node[below, text=black!60] {2\,s};
\draw[black!40] (5.38,-2.93) -- ++(0,-0.06) node[below, text=black!60] {4\,s};
\draw[black!40] (6.82,-2.93) -- ++(0,-0.06) node[below, text=black!60] {6\,s};
\draw[black!40] (8.26,-2.93) -- ++(0,-0.06) node[below, text=black!60] {8\,s};
\draw[black!40] (9.70,-2.93) -- ++(0,-0.06) node[below, text=black!60] {10\,s};
\draw[black!40] (11.14,-2.93) -- ++(0,-0.06) node[below, text=black!60] {12\,s};
\draw[black!40] (12.58,-2.93) -- ++(0,-0.06) node[below, text=black!60] {14\,s};
\draw[black!40] (14.02,-2.93) -- ++(0,-0.06) node[below, text=black!60] {16\,s};
\node[ev, fill=blue!65] (n1) at (3.94,0.00) {1};
\node[ev, fill=violet!70] (n2) at (4.66,-0.62) {2};
\node[ev, fill=orange!85] (n3) at (7.90,-1.24) {3};
\node[ev, fill=teal!70] (n4) at (3.67,-1.86) {4};
\node[ev, fill=red!70] (n5) at (3.67,-2.48) {5};
\node[ev, fill=teal!70] (n6) at (8.95,-1.86) {6};
\node[ev, fill=red!70] (n7) at (8.95,-2.48) {7};
\node[ev, fill=violet!70] (n8) at (5.38,-0.62) {8};
\node[ev, fill=blue!65] (n9) at (6.82,0.00) {9};
\node[ev, fill=blue!65] (n10) at (9.70,0.00) {10};
\node[ev, fill=violet!70] (n11) at (12.58,-0.62) {11};
\node[ev, fill=blue!65] (n12) at (12.58,0.00) {12};
\node[ev, fill=teal!70] (n13) at (13.93,-1.86) {13};
\node[ev, fill=blue!65] (n14) at (14.38,0.00) {14};
\draw[same] (n3) -- (n1);
\draw[same] (n3) -- (n5);
\draw[same] (n3) -- (n7);
\draw[sup] (n3) to (n7);
\draw[same] (n3) -- (n8);
\draw[same] (n3) -- (n9);
\draw[same] (n3) -- (n10);
\draw[same] (n3) -- (n12);
\draw[sup] (n3) to (n12);
\draw[same] (n3) -- (n14);
\draw[sup] (n3) to (n14);
\draw[sup] (n1) to (n3);
\draw[same] (n1) -- (n5);
\draw[same] (n1) -- (n7);
\draw[sup] (n1) to (n7);
\draw[same] (n1) -- (n8);
\draw[sup] (n1) to (n8);
\draw[sup] (n5) to (n3);
\draw[sup] (n5) to (n1);
\draw[same] (n5) -- (n8);
\draw[sup] (n5) to (n8);
\draw[der] (n5) to (n4);
\draw[sup] (n8) to (n3);
\draw[sup] (n8) to (n7);
\draw[same] (n8) -- (n9);
\draw[sup] (n8) to (n9);
\draw[sup] (n9) to (n3);
\draw[sup] (n9) to (n7);
\draw[sup] (n10) to (n3);
\draw[sup] (n10) to (n7);
\draw[same] (n7) -- (n8);
\draw[same] (n7) -- (n9);
\draw[same] (n7) -- (n10);
\draw[same] (n7) -- (n12);
\draw[sup] (n7) to (n12);
\draw[same] (n7) -- (n14);
\draw[sup] (n7) to (n14);
\draw[der] (n7) to (n6);
\draw[same] (2.50,-3.68) -- ++(0.6,0) node[right, text=black] {same event};
\draw[draw=green!45!black, line width=0.5pt, -{Stealth[length=1.4mm]}] (4.80,-3.68) -- ++(0.6,0) node[right, text=black] {supports};
\draw[draw=black!70, line width=0.6pt, dotted, -{Stealth[length=1.4mm]}] (6.96,-3.68) -- ++(0.6,0) node[right, text=black] {derived from (speech tag from transcript)};
\end{tikzpicture}
} \\
\midrule
\textbf{Answers} (first line, segments; score $=$ P(Yes)) \\
S1: Yes / 0-4, 10-15, score 0.986 \\
S1-ts: Yes / 0-17, score 0.992 \\
S2: Yes / 0-17, score 1.000 \\
S3 v1: No / none, score 0.126; tools: window(0-17), search(gun), done \\
S4: Yes / 0-17, score 1.000 \\
\bottomrule
\end{tabular}

\end{table*}

\subsection{Faithfulness of the cited evidence}
\label{sec:results-faith}

The systems that cite nodes cite evidence their decisions depend on (Table~\ref{tab:faith}). Hiding the nodes S3 v1 cited changes its decision on \NFaithThreeCited\% of the clips where it cited any, against \NFaithThreeRandom\% when as many random nodes are hidden; for S4 the figures are \NFaithFourCited\% and \NFaithFourRandom\%. The citations are therefore faithful, but this is a property of citing stored, identified evidence, not of the graph interface: flat text is as faithful as the graph agent. The omni model cites nothing that can be checked in this way.

\begin{table}[t]
  \centering
  \caption{Deletion faithfulness: how often the decision changes when the cited nodes are hidden, against hiding as many random uncited nodes; and the mean change of the score.}
  \label{tab:faith}
  \small
  \begin{tabular}{lrrrrr}
\toprule
System & Clips & Flips, cited (\%) & Flips, random (\%) & Score shift, cited & Score shift, random \\
\midrule
S3 v1 & 485 & 46.4 & 2.9 & 0.455 & 0.032 \\
S3 v2 & 479 & 54.9 & 1.7 & 0.540 & 0.017 \\
S4 & 505 & 49.1 & 0.6 & 0.484 & 0.005 \\
\bottomrule
\end{tabular}

\end{table}

\subsection{Cost}
\label{sec:results-cost}

Flat text needs the fewest reasoning tokens (\NSysFourTokens per clip against \NSysOneTokens for S1), but the perception agents are not free. Counting them, S4 takes about \NCostFour GPU-seconds per clip against \NCostOne for S1, roughly \NCostRatio times more (Table~\ref{tab:cost}; the stages ran on different GPUs, so the figures are indicative). The agent pipeline is cheaper only when the stored observations are reused, for example to answer several questions about the same footage or to audit a decision later.

\begin{table}[t]
  \centering
  \caption{GPU time per clip, averaged over the \NSysClips test clips.}
  \label{tab:cost}
  \small
  \begin{tabular}{lrl}
\toprule
Stage & GPU s per clip & Hardware \\
\midrule
Object detectors & 4.9 & RTX 4090/3090 \\
Video descriptions & 9.5 & mixed \\
Speech recognition & 1.6 & RTX 4090 \\
Speech tags & 0.9 & RTX PRO 4500 \\
Sound tagger & 2.6 & MacBook GPU \\
S4 reasoning & 0.9 & RTX PRO 4500, vLLM batched \\
\midrule
S4 total & 20.4 &  \\
S1 omni & 6.2 & RTX PRO 6000, one clip at a time \\
\bottomrule
\end{tabular}

\end{table}

\subsection{Two artefacts of the perception agents}
\label{sec:results-artefacts}

Two problems found while checking the graphs by hand are worth reporting for anyone using a vision-language model as a sensor. A first prompt asked the describer to mention weapons, fire, vehicles or injuries only if it saw them; it then ended \NCapNegOld\% of its descriptions with a sentence such as ``No weapons, fire, vehicles or injuries are visible'', sometimes right after describing a gun. Those sentences produced \NGraphContraOld contradicts edges; with a neutral prompt, \NCapNegNew\% of descriptions contain a negation and \NGraphEContradicts contradicts edges remain. Real contradictions between agents are rare in this domain. Similarly, a first definition of a threat that included giving orders made the speech tagger mark \NThreatOld lines as threats, many of them ordinary commands in film dialogue; a definition limited to threats of violence gave \NThreatNew. All results above use the corrected versions.

\section{Limitations}
\label{sec:limits}

\paragraph{One task, one dataset.} All results are on XD-Violence, whose clips come mostly from films, web videos and news, judged as if they were monitored footage. The violence labels serve as a proxy for the operator decision; Section~\ref{sec:human} shows the proxy holds on the annotated sample, and the limits of that sample are in Section~\ref{sec:human-limits}. In this domain the agents rarely disagree (Section~\ref{sec:results-artefacts}), which leaves little for supports and contradicts edges to do. Whether structure helps where sources do conflict, as in news verification, is not tested here.

\paragraph{The graph and its interface.} Edges were built by fixed word lists and time overlap, not learned; a learned relation model might give edges that help. The graph agent's tools, summaries and prompts are one design among many. We tried four variants on a development set under a fixed rule and report all of them, but other designs might do better. The development clips come from the training videos of the same dataset; they share no file with the test set but some come from the same films.

\paragraph{Models.} The text-based systems use one reasoning model, with one larger model and one paraphrase of the instructions as checks (Section~\ref{sec:results-robust}). S3 v2 was not rerun under either check, and the paraphrase covers the system instructions, not the task question. The single-model baseline uses one audio-visual model and one prompt, plus the time-labelled variant. Scores come from the first answer token, which saturates near 0 and 1 for these models and limits how finely clips are ranked.

\paragraph{Metrics and cost.} Our systems give one score per clip, which caps frame-level AP; frame AP and grounding are therefore secondary. Faithfulness is measured by deletion only, and cannot be measured for the single audio-visual model, which cites nothing. GPU times come from different GPUs and from batched serving for the text systems against one clip at a time for the single model, so the cost ratio is indicative.

\section{Conclusion}
\label{sec:conclusion}

We held a set of perception agents and a reasoning model fixed and varied only how their observations reach the decision about a violent clip. Reading every observation as flat text, in one pass, was as accurate as a strong audio-visual model given the raw clip, and more accurate than letting the agents exchange messages or letting the reasoning model explore an evidence graph with tools; this held with reworded prompts and with a larger reasoning model. The graph agent did not lose for lack of information: it usually saw the evidence that flat text relied on, but its tools handed it fragments, too little of the clip or detections without their context, and it often stopped early or decided against what it had read. Sound helped, as it did for the human annotators, and citations of stored observations were faithful whichever interface produced them.

For people building such pipelines, the lesson is to keep structure where it pays: as a record. Typed, timed observations with their provenance are what make citations checkable and let the same footage be questioned again without rerunning the perception models. As the interface through which the model reasons, the same structure cost accuracy on this task. Whether it helps when sources genuinely disagree, as they do in news verification, is the question we take up next.

\section*{Data and code availability}

The human reference of Section~\ref{sec:human} is available on request at \url{https://huggingface.co/datasets/amine0110/xd-violence-review} under CC BY-NC 4.0. It contains annotations only: each record names the XD-Violence clip it refers to, and users obtain the videos themselves from the official XD-Violence page under its terms. Version v1.3 is the one used in this paper. The code of all systems, the evaluation and the scripts that produce every number in this paper are at \url{https://github.com/amine0110/evidence-graph-agents}.

\section*{Ethics statement}

\paragraph{Footage.} XD-Violence consists of clips from films, web videos and news. We do not redistribute any video or frame, and we report aggregate results only. The systems decide whether a clip needs a human to look at it; they do not identify people, infer intent, or act on their own, and we make no claim about their use on real monitored footage.

\paragraph{Annotators.} Two people annotated the clips: one of the authors (A) and a volunteer (B) with no background in video analysis or security, who took no part in the study and received no payment. The first author explained the protocol and the nature of the footage to annotator~B before annotation began. Both appear in the released data only as the letters A and B. Annotator~B agreed to the release of her annotations. No formal ethics review was conducted.

\paragraph{Language models.} All models were run on the authors' own computer and on university servers; no clip or transcript was sent to an external model service.

\bibliographystyle{unsrtnat}
\bibliography{references}

@inproceedings{wu2020xdviolence,
  author    = {Wu, Peng and Liu, Jing and Shi, Yujia and Sun, Yujia and Shao, Fangtao and Wu, Zhaoyang and Yang, Zhiwei},
  title     = {Not only Look, but also Listen: Learning Multimodal Violence Detection under Weak Supervision},
  booktitle = {European Conference on Computer Vision (ECCV)},
  pages     = {322--339},
  year      = {2020},
  doi       = {10.1007/978-3-030-58577-8_20}
}

@inproceedings{gong2021ast,
  author    = {Gong, Yuan and Chung, Yu-An and Glass, James},
  title     = {{AST}: Audio Spectrogram Transformer},
  booktitle = {Proc. Interspeech},
  pages     = {571--575},
  year      = {2021},
  doi       = {10.21437/Interspeech.2021-698}
}

@inproceedings{gemmeke2017audioset,
  author    = {Gemmeke, Jort F. and Ellis, Daniel P. W. and Freedman, Dylan and Jansen, Aren and Lawrence, Wade and Moore, R. Channing and Plakal, Manoj and Ritter, Marvin},
  title     = {Audio Set: An Ontology and Human-Labeled Dataset for Audio Events},
  booktitle = {IEEE International Conference on Acoustics, Speech and Signal Processing (ICASSP)},
  pages     = {776--780},
  year      = {2017},
  doi       = {10.1109/ICASSP.2017.7952261}
}

@article{cohen1960kappa,
  author  = {Cohen, Jacob},
  title   = {A Coefficient of Agreement for Nominal Scales},
  journal = {Educational and Psychological Measurement},
  volume  = {20},
  number  = {1},
  pages   = {37--46},
  year    = {1960},
  doi     = {10.1177/001316446002000104}
}

@book{efron1993bootstrap,
  author    = {Efron, Bradley and Tibshirani, Robert J.},
  title     = {An Introduction to the Bootstrap},
  publisher = {Chapman \& Hall},
  year      = {1993}
}

@inproceedings{cheng2024yoloworld,
  title     = {{YOLO-World}: Real-Time Open-Vocabulary Object Detection},
  author    = {Cheng, Tianheng and Song, Lin and Ge, Yixiao and Liu, Wenyu and Wang, Xinggang and Shan, Ying},
  booktitle = {Proceedings of the IEEE/CVF Conference on Computer Vision and Pattern Recognition (CVPR)},
  year      = {2024}
}

@inproceedings{radford2023whisper,
  title     = {Robust Speech Recognition via Large-Scale Weak Supervision},
  author    = {Radford, Alec and Kim, Jong Wook and Xu, Tao and Brockman, Greg and McLeavey, Christine and Sutskever, Ilya},
  booktitle = {Proceedings of the 40th International Conference on Machine Learning (ICML)},
  year      = {2023}
}

@inproceedings{zanella2024lavad,
  title     = {Harnessing Large Language Models for Training-free Video Anomaly Detection},
  author    = {Zanella, Luca and Menapace, Willi and Mancini, Massimiliano and Wang, Yiming and Ricci, Elisa},
  booktitle = {Proceedings of the IEEE/CVF Conference on Computer Vision and Pattern Recognition (CVPR)},
  year      = {2024}
}

@misc{qwen3vl,
  title        = {{Qwen3-VL-8B-Instruct}},
  author       = {{Qwen Team}},
  howpublished = {\url{https://huggingface.co/Qwen/Qwen3-VL-8B-Instruct}},
  year         = {2025},
  note         = {Model card}
}

@misc{qwen3omni,
  title        = {{Qwen3-Omni-30B-A3B-Instruct}},
  author       = {{Qwen Team}},
  howpublished = {\url{https://huggingface.co/Qwen/Qwen3-Omni-30B-A3B-Instruct}},
  year         = {2025},
  note         = {Model card}
}

@misc{gemma4,
  title        = {{Gemma 4 12B} instruction-tuned},
  author       = {{Google}},
  howpublished = {\url{https://huggingface.co/google/gemma-4-12B-it}},
  year         = {2026},
  note         = {Model card}
}

@inproceedings{wang2024multiagentbounds,
  title     = {Rethinking the Bounds of {LLM} Reasoning: Are Multi-Agent Discussions the Key?},
  author    = {Wang, Qineng and Wang, Zihao and Su, Ying and Tong, Hanghang and Song, Yangqiu},
  booktitle = {Proceedings of the 62nd Annual Meeting of the Association for Computational Linguistics (Volume 1: Long Papers)},
  pages     = {6106--6131},
  year      = {2024},
  doi       = {10.18653/v1/2024.acl-long.331}
}

@article{kim2025scalingagents,
  title   = {Towards a Science of Scaling Agent Systems},
  author  = {Kim, Yubin and Gu, Ken and Park, Chanwoo and Park, Chunjong and Schmidgall, Samuel and Heydari, A. Ali and Yan, Yao and Zhang, Zhihan and Zhuang, Yuchen and Liu, Yun and Malhotra, Mark and Liang, Paul Pu and Park, Hae Won and Yang, Yuzhe and Xu, Xuhai and Du, Yilun and Patel, Shwetak and Althoff, Tim and McDuff, Daniel and Liu, Xin},
  journal = {arXiv preprint arXiv:2512.08296},
  year    = {2025}
}

@article{cemri2025mast,
  title   = {Why Do Multi-Agent {LLM} Systems Fail?},
  author  = {Cemri, Mert and Pan, Melissa Z. and Yang, Shuyi and Agrawal, Lakshya A. and Chopra, Bhavya and Tiwari, Rishabh and Keutzer, Kurt and Parameswaran, Aditya and Klein, Dan and Ramchandran, Kannan and Zaharia, Matei and Gonzalez, Joseph E. and Stoica, Ion},
  journal = {arXiv preprint arXiv:2503.13657},
  year    = {2025}
}

@article{salemi2025blackboard,
  title   = {{LLM}-Based Multi-Agent Blackboard System for Information Discovery in Data Science},
  author  = {Salemi, Alireza and Parmar, Mihir and Goyal, Palash and Song, Yiwen and Yoon, Jinsung and Zamani, Hamed and Pfister, Tomas and Palangi, Hamid},
  journal = {arXiv preprint arXiv:2510.01285},
  year    = {2025}
}

@article{han2025blackboard,
  title   = {Exploring Advanced {LLM} Multi-Agent Systems Based on Blackboard Architecture},
  author  = {Han, Bochen and Zhang, Songmao},
  journal = {arXiv preprint arXiv:2507.01701},
  year    = {2025}
}

@article{duan2026agenticvau,
  title   = {{AgenticVAU}: Multi-Agent Explore-Verify Reasoning for Video Anomaly Understanding},
  author  = {Duan, Yuxiang and Li, Huining and Li, Ao and Feng, Shuai and Kong, Lanju and Liu, Ning and Zhang, Jian and Sheng, Xingdong and Du, Yuntao},
  journal = {arXiv preprint arXiv:2608.03779},
  year    = {2026}
}

@inproceedings{yang2025panda,
  title     = {{PANDA}: Towards Generalist Video Anomaly Detection via Agentic {AI} Engineer},
  author    = {Yang, Zhiwei and Gao, Chen and Shou, Mike Zheng},
  booktitle = {Advances in Neural Information Processing Systems (NeurIPS)},
  year      = {2025}
}

@article{bekit2026qvad,
  title   = {{QVAD}: A Question-Centric Agentic Framework for Efficient and Training-Free Video Anomaly Detection},
  author  = {Bekit, Lokman and Karim, Hamza and Nguyen, Nghia T. and Yilmaz, Yasin},
  journal = {arXiv preprint arXiv:2604.03040},
  year    = {2026}
}

@inproceedings{shao2025eventvad,
  title     = {{EventVAD}: Training-Free Event-Aware Video Anomaly Detection},
  author    = {Shao, Yihua and He, Haojin and Li, Sijie and Chen, Siyu and Long, Xinwei and Zeng, Fanhu and Fan, Yuxuan and Zhang, Muyang and Yan, Ziyang and Ma, Ao and Wang, Xiaochen and Tang, Hao and Wang, Yan and Li, Shuyan},
  booktitle = {Proceedings of the 33rd ACM International Conference on Multimedia (ACM MM)},
  year      = {2025}
}

@inproceedings{wu2024vadclip,
  title     = {{VadCLIP}: Adapting Vision-Language Models for Weakly Supervised Video Anomaly Detection},
  author    = {Wu, Peng and Zhou, Xuerong and Pang, Guansong and Zhou, Lingru and Yan, Qingsen and Wang, Peng and Zhang, Yanning},
  booktitle = {Proceedings of the AAAI Conference on Artificial Intelligence},
  year      = {2024}
}

@inproceedings{zhang2025holmesvau,
  title     = {{Holmes-VAU}: Towards Long-term Video Anomaly Understanding at Any Granularity},
  author    = {Zhang, Huaxin and Xu, Xiaohao and Wang, Xiang and Zuo, Jialong and Huang, Xiaonan and Gao, Changxin and Zhang, Shanjun and Yu, Li and Sang, Nong},
  booktitle = {Proceedings of the IEEE/CVF Conference on Computer Vision and Pattern Recognition (CVPR)},
  year      = {2025}
}

@inproceedings{ghadiya2024crossmodal,
  title     = {Cross-Modal Fusion and Attention Mechanism for Weakly Supervised Video Anomaly Detection},
  author    = {Ghadiya, Ayush and Kar, Purbayan and Chudasama, Vishal and Wasnik, Pankaj},
  booktitle = {CVPR Workshop on Multimodal Learning and Applications (MULA)},
  year      = {2024}
}

@inproceedings{yang2026mars,
  title     = {Training-Free and Interpretable Hateful Video Detection via Multi-stage Adversarial Reasoning},
  author    = {Yang, Shuonan and Zhang, Yuchen and Fu, Zeyu},
  booktitle = {IEEE International Conference on Acoustics, Speech and Signal Processing (ICASSP)},
  year      = {2026}
}

@inproceedings{wang2026vto,
  title     = {{VTO}: Visual Tool Orchestration for Video Anomaly Detection},
  author    = {Wang, Rui and Wu, Yeteng and Zhang, Xianling and Qi, Mengshi},
  booktitle = {Proceedings of the 34th ACM International Conference on Multimedia (ACM MM)},
  year      = {2026}
}

@article{wu2025avadclip,
  title   = {{AVadCLIP}: Audio-Visual Collaboration for Robust Video Anomaly Detection},
  author  = {Wu, Peng and Su, Wanshun and Pang, Guansong and Sun, Yujia and Yan, Qingsen and Wang, Peng and Zhang, Yanning},
  journal = {arXiv preprint arXiv:2504.04495},
  year    = {2025}
}

@misc{qwen36,
  title        = {{Qwen3.6-35B-A3B}},
  author       = {{Qwen Team}},
  howpublished = {\url{https://huggingface.co/Qwen/Qwen3.6-35B-A3B}},
  year         = {2026},
  note         = {Model card}
}

\appendix
\section{Prompts}
\label{app:prompts}

All prompts are exported from the code that ran them (the two replaced versions are kept verbatim). In S3 the overview of Section~\ref{sec:method-systems} follows the system prompt; in S4 the nodes follow it, one line each, in the format of Table~\ref{tab:example}. Every system ends with the task question; S2, S3 and S4 add the citation line.

\paragraph{Task question (all systems; the final turn of S2, S3 and S4).}
\begin{quote}\footnotesize\ttfamily
You are assisting a security operator who monitors video. Decide whether this clip needs the operator's attention because it shows violence or danger: fighting, shooting, a riot, abuse, a car accident or an explosion. Films and news footage are included; judge what is shown, not whether it is staged.\\
Answer on the first line with Yes or No only. On the second line, give the time segments in seconds where it happens, like 12-20, 31-40, or write none.
\end{quote}

\paragraph{Line added to the final turn of S2, S3 and S4.}
\begin{quote}\footnotesize\ttfamily
On a third line, list the ids of the evidence items your answer rests on, like n3, n17, or write none.
\end{quote}

\paragraph{S4, system prompt (the nodes follow, one per line).}
\begin{quote}\footnotesize\ttfamily
You are the reasoning agent of a video review system. Perception agents have processed one video clip; below is everything they found, sorted by time: video descriptions (one per 4 seconds), object detections, sound events, speech transcripts, and speech tagged as threat, distress or report of violence. Detectors and descriptions can be wrong; weigh them against each other.
\end{quote}

\paragraph{S4, paraphrased system prompt (prompt-sensitivity check).}
\begin{quote}\footnotesize\ttfamily
Your job is to judge one video clip for a review system. Before you, several perception models went through the clip; their findings are listed below in time order: a description of every 4 seconds of video, object detections, sound events, transcribed speech, and speech marked as a threat, as distress, or as a report of violence. Any of these models can be wrong, so compare their findings before deciding.
\end{quote}

\paragraph{S3 v1, system prompt (the overview follows).}
\begin{quote}\footnotesize\ttfamily
You are the reasoning agent of a video review system. Perception agents have already processed one video clip and stored what they found as evidence items in a shared evidence graph: video descriptions (one per 4 seconds), object detections, sound events, speech transcripts, and speech tagged as threat, distress or report of violence. Items carry a time span and, where available, a confidence. Links between items say whether two items are the same event, support each other, or contradict each other. Detectors and descriptions can be wrong; weigh them against each other.\\
~\\
You read the graph with tools. Reply with exactly one JSON object per turn, nothing else:\\
\{"tool": "window", "start": 10, "end": 30\}            items overlapping a time span\\
\{"tool": "window", "start": 10, "end": 30, "kind": "caption"\}   same, one kind only (caption, detection, sound\_event, transcript\_segment, speech\_tag)\\
\{"tool": "search", "concept": "gun"\}                  items asserting a concept: gun, knife, fire, fight, crash, injury, distress, threat, emergency, crowd, vehicle\\
\{"tool": "links", "id": "n12"\}                        items linked to n12 (same\_event, supports, contradicts)\\
\{"tool": "provenance", "id": "n12"\}                   which agent and model produced n12\\
\{"tool": "done"\}                                      you have enough evidence to answer\\
You have at most 8 tool calls.
\end{quote}

\paragraph{S3 v1, paraphrased opening paragraph (prompt-sensitivity check).}
\begin{quote}\footnotesize\ttfamily
Your job is to judge one video clip for a review system. Before you, several perception models went through the clip and saved their findings in a shared evidence graph: a description of every 4 seconds of video, object detections, sound events, transcribed speech, and speech marked as a threat, as distress, or as a report of violence. Each finding has a time span and, when one exists, a confidence. Links tell you when two findings describe the same event, back each other up, or contradict each other. Any of these models can be wrong, so compare their findings before deciding.
\end{quote}

\paragraph{S3 v2, system prompt.}
\begin{quote}\footnotesize\ttfamily
You are the reasoning agent of a video review system. Perception agents have already processed one video clip and stored what they found as evidence items in a shared evidence graph: video descriptions (one per 4 seconds), object detections, sound events, speech transcripts, and speech tagged as threat, distress or report of violence. Items carry a time span and, where available, a confidence. Links say whether items from different agents support each other. Detectors and descriptions can be wrong: an object detector alone is weak evidence, a description or several agents agreeing is stronger.\\
~\\
You start with the video descriptions of the whole clip, a timeline of what each agent found, and the corroborated events. The object detector is often wrong on films (it mistakes phones, cameras and tools for guns), so the timeline only shows its findings where another agent agrees. Use the tools to check the sound, speech and detections behind the flagged spans and events before answering.\\
~\\
Reply with exactly one JSON object per turn, nothing else:\\
\{"tool": "window", "start": 10, "end": 30\}            items overlapping a time span\\
\{"tool": "window", "start": 10, "end": 30, "kind": "caption"\}   same, one kind only (caption, detection, sound\_event, transcript\_segment, speech\_tag)\\
\{"tool": "event", "n": 2\}                             all items of corroborated event 2\\
\{"tool": "search", "concept": "gun"\}                  items asserting a concept: gun, knife, fire, fight, crash, injury, distress, threat, emergency, crowd, vehicle\\
\{"tool": "links", "id": "n12"\}                        items linked to n12 (same\_event, supports, contradicts)\\
\{"tool": "done"\}                                      you have read enough to answer\\
You have at most 12 tool calls.
\end{quote}

\paragraph{S2, reasoning agent.}
\begin{quote}\footnotesize\ttfamily
You are the reasoning agent of a video review system. Three perception agents processed one video clip and each sent you a report: video, sound and speech. Their tools can be wrong. You may ask an agent a follow-up question. Reply with exactly one JSON object per turn, nothing else:\\
\{"ask": "video", "question": "..."\}      ask an agent (video, audio\_events or speech)\\
\{"done": true\}                           you have enough to answer\\
You may ask at most 3 questions.
\end{quote}

\paragraph{S2, request sent to each perception agent.}
\begin{quote}\footnotesize\ttfamily
Write a short report (at most 120 words) for the reasoning agent, who must decide whether the clip needs a security operator's attention. Say what you found and when (in seconds), and how sure you are. Do not invent anything that is not in your data.
\end{quote}

\paragraph{Speech tagger (version used).}
\begin{quote}\footnotesize\ttfamily
You read speech transcribed automatically from a short video clip; the transcript may contain errors and may be in any language. Classify the TARGET line only; the lines around it are context.\\
~\\
A: threat of violence: the speaker threatens to hurt or kill someone, or demands something under threat of violence. Ordinary orders, arguments, insults and rude remarks are not threats.\\
B: distress: the speaker screams, begs for help or mercy, or expresses pain or terror. Ordinary surprise, sadness or annoyance is not distress.\\
C: report of violence: the speaker describes violence, a weapon, an injury, an accident or an explosion.\\
D: none of these.\\
~\\
Before:\\
\{before\}\\
TARGET [\{start:.1f\}-\{end:.1f\} s]: \{text\}\\
After:\\
\{after\}\\
~\\
Answer with one letter: A, B, C or D.
\end{quote}

\paragraph{Speech tagger, first version of the two labels that changed.}
\begin{quote}\footnotesize\ttfamily
A: threat or aggression: the speaker threatens someone, gives orders at weapon point, or verbally attacks someone.\\
B: distress: the speaker screams, begs, calls for help, or expresses pain or fear.
\end{quote}

\paragraph{Video describer (version used).}
\begin{quote}\footnotesize\ttfamily
These are \{n\} frames, one second apart, from a video. In at most 40 words, describe factually what the people do and what happens. Describe only what you see; do not list things that are absent.
\end{quote}

\paragraph{Video describer, first version.}
\begin{quote}\footnotesize\ttfamily
These are \{n\} frames, one second apart, from a video. In at most 40 words, describe factually what the people do and what happens. Mention weapons, fire, vehicles or injuries only if you see them.
\end{quote}

\paragraph{S2, agent roles.} Each perception agent is told ``You are the video agent: you hold descriptions of the clip every 4 seconds and object detections.'', ``You are the sound agent: you hold the sound events an audio tagger found.'', ``You are the speech agent: you hold the transcript and the speech tagged as threat, distress or report.''

\end{document}